\documentclass[10pt,twocolumn,letterpaper]{article}

\usepackage{cvpr}              % To produce the CAMERA-READY version
\usepackage{multirow}
\usepackage{subcaption}
\usepackage{xspace}
\usepackage{algorithm}
\usepackage{algorithmic}
\usepackage{listings}
\usepackage{CJK}
\definecolor{cvprblue}{rgb}{0.21,0.49,0.74}
\usepackage[pagebackref,breaklinks,colorlinks,allcolors=cvprblue]{hyperref}
\newcommand{\model}{CogniEdit\xspace}
\def\paperID{} % *** Enter the Paper ID here
\def\confName{CVPR}
\def\confYear{2026}

\title{\model: Dense Gradient Flow Optimization for Fine-Grained Image Editing}

\author{Yan Li$^{\dagger \S}$
\quad \quad
Lin Liu$^{\ddagger}$ 
\quad \quad
Xiaopeng Zhang$^{\ddagger}$ 
\quad \quad 
Wei Xue$^{\dagger}$  
\quad 
\\
Wenhan Luo$^{\dagger}$
 \quad \quad
Yike Guo$^{\dagger}$ 
\quad \quad
Qi Tian$^{\ddagger}$
\\
{$^{\dagger}$ Hongkong University of Science and Technology}
\quad 
{$^{\ddagger}$ Huawei Company}
}

\begin{document}
\maketitle
\begingroup
\renewcommand{\thefootnote}{\fnsymbol{footnote}}
\footnotetext[4]{For any inquiries, please reach out to: ylitz@connect.ust.hk}
\endgroup

\begin{abstract}
%Instruction-based image editing has made significant progress with diffusion models, yet existing methods struggle to accurately follow fine-grained instructions that specify detailed attributes such as precise colors, positions, and quantities. While recent approaches employ reinforcement learning techniques like Group Relative Policy Optimization (GRPO) to align models with human preferences, they typically apply optimization only at the final generation step, providing sparse feedback that limits fine-grained control. In this work, we propose a novel dense reward optimization strategy that propagates gradients at each sampling step during the diffusion process, enabling more precise alignment with detailed editing instructions. Our approach consists of three key components: (1) a dynamic token focus relocation mechanism that adaptively emphasizes semantically important terms in editing instructions, (2) a dense GRPO optimization strategy that provides continuous feedback throughout the generation process, and (3) an efficient training mechanism that accumulates gradients over consecutive sampling steps. Extensive experiments on benchmark datasets demonstrate that our method achieves state-of-the-art performance in fine-grained instruction following, attaining the highest visual quality scores across all evaluation categories while preserving general editing capabilities. Our approach successfully addresses the fundamental challenge of balancing high-quality visual generation with faithful adherence to detailed editing instructions.
Instruction-based image editing with diffusion models has achieved impressive results, yet existing methods struggle with fine-grained instructions specifying precise attributes such as colors, positions, and quantities. While recent approaches employ Group Relative Policy Optimization (GRPO) for alignment, they optimize only at individual sampling steps, providing sparse feedback that limits trajectory-level control. We propose a unified framework \textbf{\model}, combining multi-modal reasoning with dense reward optimization that propagates gradients across consecutive denoising steps, enabling trajectory-level gradient flow through the sampling process. Our method comprises three components: (1) Multi-modal Large Language Models for decomposing complex instructions into actionable directives, (2) Dynamic Token Focus Relocation that adaptively emphasizes fine-grained attributes, and (3) Dense GRPO-based optimization that propagates gradients across consecutive steps for trajectory-level supervision. Extensive experiments on benchmark datasets demonstrate that our \model achieves state-of-the-art performance in balancing fine-grained instruction following with visual quality and editability preservation. %Extensive experiments demonstrate state-of-the-art performance: we achieve the highest visual quality scores across all domains on Kris-Bench while maintaining competitive performance on general editing benchmarks, successfully balancing fine-grained instruction following with visual quality and editability preservation.
\end{abstract}    
\section{Introduction}
\label{sec:intro}

\begin{figure}[h]
    \centering
     \vspace{-8pt}
    \includegraphics[width=\linewidth]{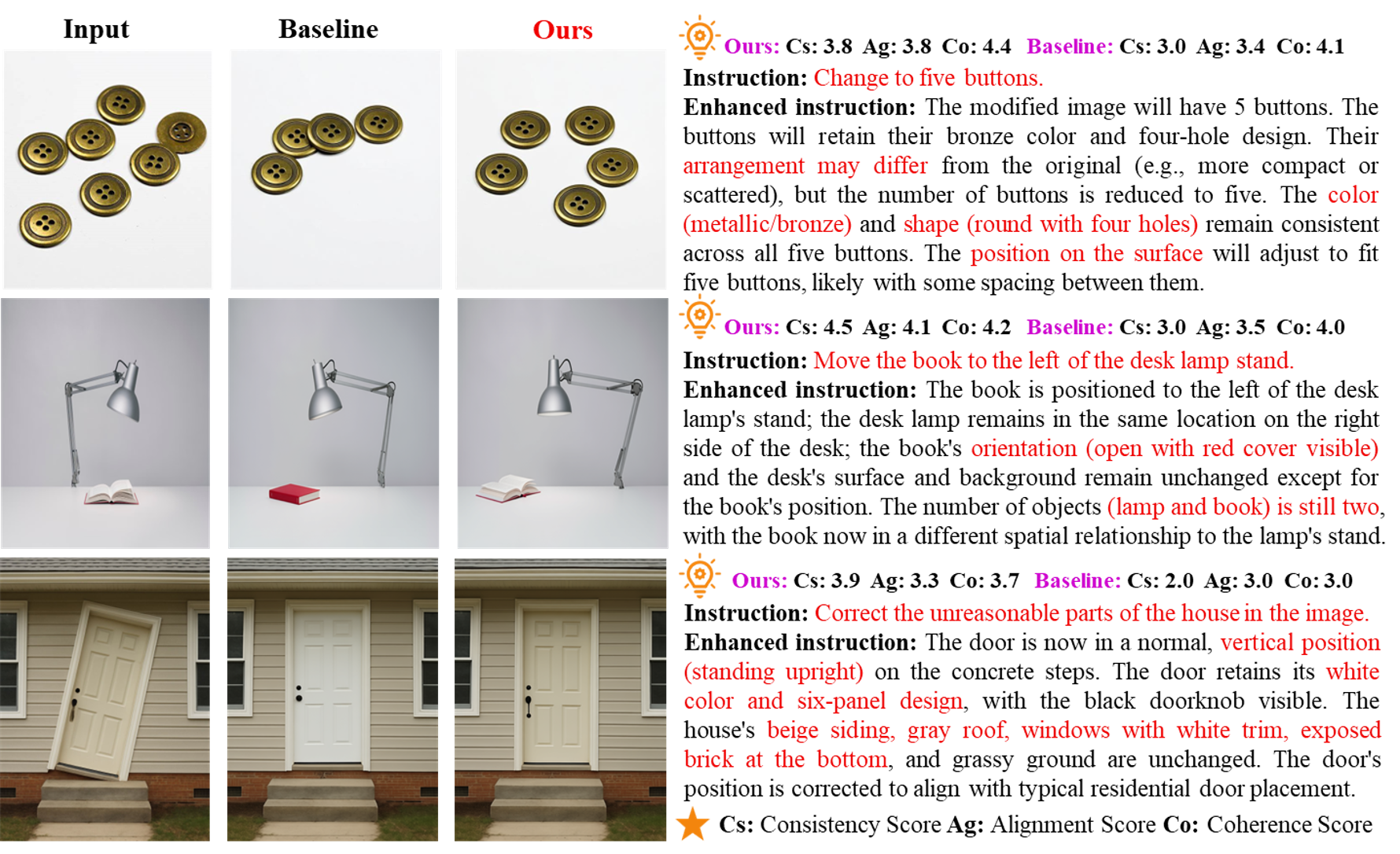}
    \caption{Comparison of editing results on instructions with fine-grained numerical and spatial attributes. We highlight three evaluation metrics: \textbf{Consistency} (visual consistency between edited and input images), \textbf{Coherence} (logical plausibility of edited content), and \textbf{Alignment} (instruction-image alignment), evaluated by Qwen2.5-VL-Think (7B) model.}
    \label{fig:front_img}
    \vspace{-12pt}
\end{figure}

Instruction-based image editing enables users to manipulate images through natural language commands, offering an intuitive interface for visual content creation. Recent advances in diffusion models~\cite{peebles2023scalable,lu2024hierarchical} have achieved impressive results in generating high-quality edits that align with user instructions~\cite{brooks2023instructpix2pix,zhang2023magicbrush,wu2025qwen,liu2025step1x}. However, critical limitations persist: existing methods~\cite{yu2025anyedit,xiao2025omnigen,huang2024smartedit} struggle with both fine-grained instruction following and reasoning-intensive editing tasks. They fail to accurately capture precise visual attributes such as exact colors, specific positions, or numerical quantities. Moreover, when faced with complex instructions requiring reasoning or domain knowledge, these models exhibit an understanding gap between instruction semantics and editing demands~\cite{wu2025kris}, leading to edits that are semantically incorrect or factually implausible. These limitations stem from the training paradigm: most instruction-based editing models~\cite{brooks2023instructpix2pix,zhang2023magicbrush} are trained through supervised learning on paired datasets, which primarily optimize for overall visual similarity between generated and target images. While this enables general editing capabilities, it does not explicitly optimize for fine-grained alignment between textual descriptions and specific visual attributes. As illustrated in Figure~\ref{fig:front_img}, existing methods struggle with numerical and spatial attributes, whereas our approach achieves superior alignment between instructions and visual attributes.% while maintaining consistency with the source image.

Recently, reinforcement learning (RL) techniques have shown promise in aligning generative models with human preferences~\cite{black2023training,fan2023reinforcement,wallace2024diffusion,clark2023directly,lee2023aligning}. Group Relative Policy Optimization (GRPO)~\cite{shao2024deepseekmath}, in particular, has demonstrated success in text generation~\cite{wang2025grpo,li2025optimizing} and image generation~\cite{xue2025dancegrpo,wang2025pref,liu2025flow} by computing advantages relative to group statistics without requiring separate value networks. However, existing GRPO applications~\cite{xue2025dancegrpo,liu2025flow} typically optimize each sampling step independently, treating each step as a separate decision point. This single-step optimization strategy fails to guide the overall sampling process while the model learns to improve individual steps, gradients do not flow through consecutive sampling steps, preventing dense supervision across the denoising process. For image editing, where consistency between source and edited regions is paramount~\cite{wang2023instructedit,li2024zone}, this lack of gradient flow hinders fine-grained alignment and can lead to visual artifacts or convergence difficulties~\cite{wu2024deep}.  Achieving precise control over editing attributes while maintaining source consistency requires both fine-grained alignment between text features and visual content, and dense supervision across the sampling process to ensure edits evolve correctly from coarse structure to fine details.

%Achieving precise control over specific editing attributes while maintaining source consistency requires dense supervision that guides the sampling direction at each denoising step, ensuring edits evolve correctly from coarse structure to fine details.

%To address these challenges, we propose a unified framework that combines multi-modal reasoning with dense reward optimization for instruction-based image editing. Our key insight is twofold:  one is reasoning-intensive editing tasks require semantic understanding beyond visual pattern matching, and the other is fine-grained instruction following demands dense supervision throughout the diffusion process. We first leverage Multi-modal Large Language Models (MLLMs) to bridge the understanding gap, decomposing complex or knowledge-intensive instructions into clear, actionable editing directives that ensure semantic correctness and factual plausibility. For precise execution, we introduce two synergistic components: (i) \textit{Dynamic Token Focus Relocation}, which adaptively emphasizes semantically important terms specifying fine-grained attributes in the reasoned instructions, and (ii) \textit{Dense GRPO Optimization}, which accumulates gradients across consecutive denoising steps to enable gradient flow through the sampling trajectory while maintaining computational efficiency. By unifying MLLM-based semantic reasoning with trajectory-aware dense optimization, our method achieves both complex reasoning-intensive editing and precise fine-grained attribute control while naturally preserving visual coherence.
To address these challenges, we propose a unified framework \textbf{\model}, combining multi-modal reasoning with dense reward optimization for instruction-based image editing. Our key insight is twofold: first, reasoning-intensive editing requires semantic understanding beyond visual pattern matching; second, fine-grained instruction following demands dense supervision throughout the editing process. We leverage Multi-modal Large Language Models (MLLMs) to decompose complex instructions into clear, actionable editing directives ensuring semantic correctness. For precise execution, we introduce two synergistic components: (i) \textit{Dynamic Token Focus Relocation}, which adaptively emphasizes fine-grained attributes in reasoned instructions, and (ii) \textit{Dense GRPO-based optimization}, which propagates gradients across consecutive denoising steps to enable gradient flow through the sampling trajectory. By unifying MLLM-based reasoning with trajectory-aware dense optimization, our method achieves both complex reasoning-intensive editing and precise fine-grained attribute control.

Our contributions can be summarized as follows:
\begin{itemize}
    \item We address two critical limitations in existing image editing methods: understanding gap in reasoning-intensive tasks and sparse feedback in optimization by proposing a unified framework combining MLLM-based reasoning with dense reward optimization.
    \item We introduce a dynamic token focus relocation mechanism that enables adaptive attention to fine-grained attribute specifications in editing instructions, enhancing the model's ability to capture precise visual details.
    \item We develop an efficient dense GRPO-based optimization mechanism with gradient propagates through sampling trajectory, achieving superior fine-grained instruction following while maintaining computational efficiency and training stability.
\end{itemize}
\section{Related Work}
\label{sec:formatting}
\subsection{General Image Editing}
%\subsection{Instruction-based Image Editing}
\textbf{Instruction-based Image Editing.}
Instruction-based image editing enables users to manipulate images through natural language commands. Existing approaches can be categorized into training-free and training-based methods. Training-free methods~\cite{zhu2025kv,hertz2022prompt,cao2023masactrl,han2024proxedit,patashnik2023localizing,li2024zone,huang2023region,avrahami2025stable} manipulate attention mechanisms or incorporate guidance during diffusion, offering flexibility without model updates but often struggling with complex semantic edits. Training-based methods~\cite{batifol2025flux,brooks2023instructpix2pix,zhang2023magicbrush,yu2025anyedit,huang2024smartedit} learn from paired datasets of source images, instructions, and targets. InstructPix2Pix~\cite{brooks2023instructpix2pix} pioneered this direction, while recent large-scale models such as Qwen-Image~\cite{wu2025qwen} and OmniGen~\cite{wu2025omnigen2,xiao2025omnigen} achieve impressive general-purpose capabilities through massive pre-training.
However, despite these advances, existing methods struggle with fine-grained instruction following, particularly when instructions require reasoning or domain knowledge. This limitation results in misalignment between textual instructions and visual content, motivating the integration of external knowledge and reasoning capabilities.

\textbf{Knowledge-enhanced Image Editing.} 
To handle instructions requiring reasoning or domain knowledge, recent methods leverage Multi-modal Large Language Models (MLLMs) to enhance semantic understanding~\cite{liu2025step1x,li2024brushedit,wang2023instructedit,yeh2025beyond,fu2023guiding,wang2024genartist}. Step1X-Edit~\cite{liu2025step1x} and BrushEdit~\cite{li2024brushedit} employ MLLMs to jointly process images and instructions for reasoning about edit requirements. GenArtist~\cite{wang2024genartist} uses GPT-4~\cite{achiam2023gpt} as an agent to decompose complex tasks into simpler instructions. While these approaches improve semantic accuracy through better instruction understanding, they primarily focus on preprocessing instructions before generation. The generation process itself remains supervised learning on paired data, which does not explicitly optimize for fine-grained alignment between textual specifications and visual attributes.

In contrast, our method addresses fine-grained alignment through the optimization process itself. We combine MLLM-based reasoning with dynamic token focus relocation to enhance the model's ability to capture fine-grained textual features from editing instructions, and dense reward optimization that provides supervision across consecutive sampling steps. This enables our method to achieve superior fine-grained instruction following and precise alignment between textual specifications and visual attributes.

%In contrast, our method addresses fine-grained alignment through the optimization process itself. We combine MLLM-based reasoning with dense reward optimization that provides trajectory-level supervision across consecutive denoising steps. This enables gradient flow through the sampling process, allowing the model to learn precise alignment between textual specifications and visual attributes beyond what supervised learning achieves. %Our method takes a different approach by proposing a dense reward optimization to apply dense supervision across the denoising process to guide the model toward edits that are both visually appealing and semantically accurate, addressing the knowledge-grounding challenge through optimization.

\subsection{Reinforcement Learning for Image Editing}
Reinforcement learning has emerged as a powerful approach for aligning generative models with human preferences and complex objectives~\cite{black2023training,fan2023reinforcement,wallace2024diffusion,clark2023directly,lee2023aligning}. Group Relative Policy Optimization (GRPO)~\cite{shao2024deepseekmath} offers an efficient RL framework that computes advantages relative to group statistics without requiring separate value networks~\cite{wang2025pref,xue2025dancegrpo,liu2025flow}. Recent work has begun adapting these techniques to image editing, focusing on improving instruction adherence and visual quality~\cite{zhang2024hive,kumari2025learning}. However, existing approaches typically apply optimization at individual sampling steps, treating each step as an independent decision point. This sparse feedback strategy limits the model's ability to learn trajectory-level guidance across the denoising process, hindering fine-grained control over specific editing attributes~\cite{wu2024deep,fan2023reinforcement,liu2024improving}. %While dense reward strategies have been explored in generation tasks~\cite{black2023training,fan2023reinforcement,liu2024improving}, they remain underexplored for editing, where maintaining source image coherence and precise attribute control pose unique challenges.

Our \model extends standard GRPO to dense reward optimization specifically designed for image editing. By accumulating gradients across consecutive sampling steps, we enable gradient flow through the denoising trajectory, providing dense supervision throughout the editing process. This trajectory-level optimization allows the model to learn fine-grained alignment between textual specifications and visual attributes while maintaining computational efficiency and training stability.

\section{Methodology} \label{sec:method}
\begin{figure*}[h]
    \centering
    \includegraphics[width=\linewidth]{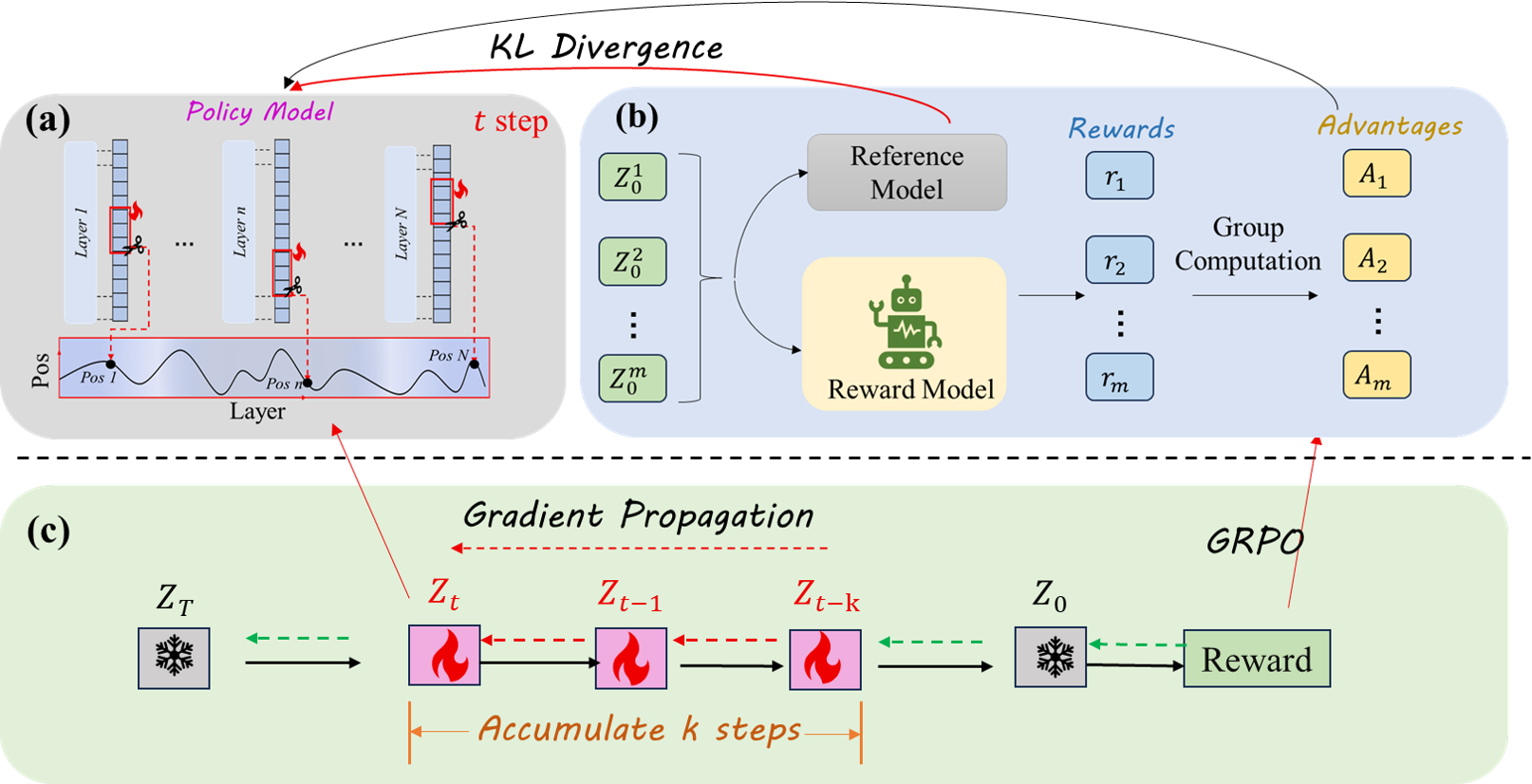}
    \caption{Overview of our proposed \model. (a) Dynamic token focus relocation mechanism for capturing fine-grained textual features. (b) Dense GRPO-based optimization strategy with gradient propagation at each sampling step. (c) Training process showing gradient accumulation over $k$ consecutive steps.}
    \label{fig:main_framework}
\end{figure*}

%We present an overview of our method in Figure~\ref{fig:main_framework}. Our approach consists of three key components: (a) a dynamic token focus relocation mechanism that enables the policy model to capture fine-grained textual features from editing instructions, (b) a dense GRPO-based optimization strategy that propagates gradients at each sampling step during the diffusion process, and (c) a training mechanism that randomly selects the $t$-th sampling step and accumulates gradients over the subsequent $k$ steps to update model parameters. These components work synergistically to enhance fine-grained instruction following while maintaining efficient optimization.
Figure~\ref{fig:main_framework} presents an overview of our \model, which consists of three key components: (a)Dynamic Token Focus Relocation that adaptively emphasizes fine-grained attributes in instructions, (b) GRPO-based optimization adapted for image editing with batch-level advantage computation, and (c) Dense GRPO-based Optimization Supervision that propagates gradients across consecutive denoising steps to enable trajectory-level gradient flow. %These components synergistically enhance fine-grained instruction following while maintaining efficient optimization.

\subsection{GRPO for Image Editing} \label{sec:grpo_for_image_editing}
Group Relative Policy Optimization (GRPO) is a reinforcement learning technique designed to align generative models with human preferences through reward-based optimization. Unlike traditional policy gradient methods that require explicit value function estimation, GRPO operates by sampling multiple outputs from the current policy and using their relative rewards to compute policy gradients. 

\textbf{Standard Flow-Matching GRPO.}
For image generation with flow-matching models, GRPO adapts policy optimization to the continuous-time diffusion framework. Let $v_\theta$ denote the velocity network parameterized by $\theta$, which predicts the direction of flow at each timestep. Given an input image and editing instruction $c$, GRPO samples a group of $G$ generated images $\{x_0^1, x_0^2, \ldots, x_0^G\}$ through the denoising process with $T$ timesteps, where each sampling trajectory is determined by the policy $\pi_\theta$ induced by the velocity network. For each generated image $x_0^i$, we compute the reward $R(x_0^i, c)$ using a reward model. The advantage for the $i$-th sample is computed relative to the group statistics:
\begin{equation}
\hat{A}^i = \frac{R(x_0^i, c) - \frac{1}{G}\sum_{j=1}^G R(x_0^j, c)}{\text{std}\left(\{ R(x_0^j, c) \}_{j=1}^G \right)}
\end{equation}
where normalization by standard deviation stabilizes training. The policy is updated by maximizing the clipped objective, $clip$ denotes the clip operation:
\begin{equation}
\begin{split}
\mathcal{J}&_{{\text{GRPO}}}(\theta) = \mathbb{E}_{\{c, x^i\}} \left[ \frac{1}{G} \sum_{i=1}^G \frac{1}{T} \sum_{t=0}^{T-1} \min \left( r_i^t(\theta) \hat{A}^i, \right. \right.   \\
~&\left. \left.  \text{clip}(r_i^t(\theta), 1-\epsilon, 1+\epsilon) \hat{A}^i \right) - \hat{\beta} D_{\text{KL}}(\pi_\theta \parallel \pi_{\text{ref}}) \right] 
\end{split}
\end{equation}
where $r_i^t(\theta) = \frac{p_\theta(x_{t-1}^i | x_t^i, c)}{p_{\theta_{\text{old}}}(x_{t-1}^i | x_t^i, c)}$ is the probability ratio at timestep $t$ for the $i$-th sample, $\epsilon$ controls the clipping range to prevent large policy updates, and $\hat{\beta}$ balances the KL divergence term for regularization against a reference policy $\pi_{\text{ref}}$. The average over both group size $G$ and timesteps $T$ ensures stable gradient estimates. However, this formulation treats each timestep independently, limiting trajectory-level guidance.

\textbf{Adapting GRPO for Image Editing.}
Image editing poses unique challenges compared to generation tasks. While generation tasks benefit from stochastic sampling to produce diverse outputs, editing tasks require high consistency and coherence with both the input image and editing instruction. Standard flow-matching models for editing use deterministic ODEs to ensure reproducible, consistent edits. However, this deterministic formulation presents a challenge for GRPO: sampling multiple outputs $\{x_0^i\}_{i=1}^G$ from a deterministic trajectory yields identical results, eliminating the diversity necessary for meaningful advantage computation.

To address this, we make two key adaptations. First, following~\cite{liu2025flow}, we inject controlled stochasticity into the sampling process by converting the deterministic ODE to an SDE via Euler-Maruyama discretization:
\begin{equation}
\begin{aligned}
x_{t + \Delta t} = x_t + s_\theta (x_t, t) \Delta t 
%( v_\theta(x_t, t) + \frac{\sigma_t^2}{2} [ x_t + \\ (1 - t)v_\theta(x_t, t)]) \Delta t 
+ \sqrt{\Delta t}\sigma_t\epsilon
\end{aligned}
\end{equation}
where $\epsilon \sim \mathcal{N}(0, 1)$, $s_\theta(x_t,t) = ( v_\theta(x_t, t) + \sigma_t^2/{2} [ x_t + (1 - t)v_\theta(x_t, t)])$ and $\sigma_t $ controls the noise level. This stochastic injection enables diverse sampling while maintaining editing quality through careful noise scheduling.
Second, we compute advantages at the batch level rather than per-instance. Since editing outputs from the same input exhibit limited diversity even with stochastic sampling, per-instance advantage normalization produces high-variance gradients. By computing statistics across a batch of $B$ editing instances, we obtain more stable advantage estimates:
$\hat{A}^{b} = \frac{1}{G}\sum_{i=1}^{G}\frac{R(x_0^{b,i}, c_b) - \mu_{\text{batch}}}{\sigma_{\text{batch}}},$ \text{where} $\quad \mu_{\text{batch}} = \frac{1}{BG}\sum_{b=1}^B\sum_{i=1}^G R(x_0^{b,i}, c_b), \quad \sigma_{\text{batch}} = \text{std}(\{R(x_0^{b,i}, c_b)\}_{b,i})$.
The editing objective becomes:
\begin{equation}
\small
\begin{split}
\mathcal{J}&_{\text{GRPO}}^{edit}(\theta) = \frac{1}{B} \sum_{b=1}^B \mathbb{E}_{\{c_b, x_0^{b}\}} \left[ \frac{1}{T} \sum_{t=0}^{T-1} \min \left( r_{b}^t(\theta)\hat{A}^{b},  \right. \right. \\
& \left. \left.   \, \text{clip}(r_{b}^t(\theta), 1-\epsilon, 1+\epsilon) \hat{A}^{b} \right) - \hat{\beta} D_{\text{KL}}(\pi_\theta \parallel \pi_{\text{ref}}) \right]
\end{split}
\end{equation}
where the superscript $b$ indexes the batch and $i$ indexes samples within each group. This batch-level normalization provides stable training dynamics while preserving the fine-grained control necessary for precise editing.

\subsection{Dynamic Token Focus Relocation} 
%A key challenge in instruction-based editing is ensuring the model attends to semantically relevant instruction tokens at each processing stage. We observe that the issue comes from the suboptimal focus patterns of the model (as analyzed in Section~\ref{sec:analy_dense}), for instance, during the attention layers, it may persistently attend to generic action words like ``change'' or ``add'' throughout the network hierarchy, while failing to adequately process specific semantic details (i.e., ``purple'', ``red'', ``five'', ``on the left'', ``in the background'', .etc.) that are essential for accurate editing. And this issue becomes more pronounced in complex instructions where different semantic components require emphasis at different stages of the generation process. Intuitively, early layers should focus on high-level semantic understanding, while deeper layers need to attend to fine-grained attributes that determine the specific visual modifications.

%To address this limitation, we propose a Dynamic Text Alignment mechanism that learns to adaptively redirect attention to semantically important tokens in a layer-specific manner.The core insight is that different network layers process different levels of semantic abstraction, and therefore should attend to different parts of the instruction. Rather than relying on fixed attention patterns, our approach dynamically predicts which tokens are most relevant at each layer and uses learnable soft tokens to guide the model's attention toward these positions.
A key challenge in instruction-based editing is ensuring the model attends to semantically relevant tokens at each processing stage. We observe that models often persistently attend to generic action words (e.g., ``change'', ``add'') throughout the network hierarchy while failing to process specific semantic details (e.g., ``purple'', ``five'', ``on the left'') essential for accurate editing (as analyzed in Section~\ref{sec:analy_dense}). This issue becomes pronounced in complex instructions where different semantic components require emphasis at different stages—intuitively, early layers should focus on high-level semantics while deeper layers attend to fine-grained attributes. To address this, we propose Dynamic Token Focus Relocation that adaptively redirects attention to semantically important tokens in a layer-specific manner. Rather than relying on fixed attention patterns, our approach dynamically predicts which tokens are most relevant at different layers and uses learnable soft tokens to guide the model's attention toward these positions, enabling different network layers to process different levels of semantic abstraction.

Formally, let $h_{i}^{l} \in \mathbb{R}^{l \times d}$ denote the encoded instruction tokens at layer $i$, where $l$ is the sequence length and $d$ is the embedding dimension. We introduce a lightweight predictor $p_{\eta}$ to identify the optimal attention focus position:
\begin{equation}
    \text{pos} = p_{\eta}(h_{i}^{l}), \quad \text{pos} \in [0, l-\xi]
\end{equation}
where $k$ denotes the number of tokens to emphasize. The predictor is trained end-to-end to learn layer-specific attention patterns. Once the position is predicted, we enforce attention toward these positions using learnable soft tokens $s_{i}^{1:\xi} = E(i) \in \mathbb{R}^{\xi \times d}$ for each layer $i$, where $E$ is a layer-dependent embedding function. These soft tokens serve as attention anchors and are injected into the instruction sequence at the predicted positions, replacing the original embeddings:
\begin{equation}
    h_{i}^{\text{pos}:\text{pos}+\xi} \leftarrow s_{i}^{1:\xi}, \quad a_{i}^{\text{pos}:\text{pos}+\xi} = s_{i}^{1:\xi} \cdot A^{l}
\end{equation}
where $A^{l} \in \mathbb{R}^{l \times l}$ is the attention map at layer $l$. The left equation replaces token embeddings, while the right equation computes attention scores using the injected soft tokens. Through joint optimization of the position predictor $p_{\eta}$ and soft tokens $\{s_{i}^{1:\xi}\}_{i=1}^{N}$ across all $N$ layers, the mechanism learns to dynamically emphasize different instruction components at appropriate depths. This results in hierarchical attention patterns where early layers focus on high-level semantics while deeper layers concentrate on fine-grained attributes, thereby improving instruction-image alignment. 

\subsection{Dense GRPO-based Optimization Supervision} \label{sec:dense_grpo}
In section~\ref{sec:grpo_for_image_editing}, we introduce the how to adopt the GRPO-based optimization strategy for the image editing tasks. However, the standard GRPO-based optimization strategy optimized at single sampling step, and provide sparse feedback to the generation process. %This approach overlooks valuable learning signals from the generation trajectory.
In diffusion-based image editing, each denoising step progressively refines the image, and the quality of these intermediate steps impacts the final result.
A model that makes poor sampling decisions early in the generation process may struggle to recover in later steps, even with correct final-step optimization.

We propose Dense GRPO-based Optimization Supervision, which extends the standard GRPO objective to provide supervision at multiple intermediate sampling steps rather than only optimized at single sampling step. And to make the gradient propagation across the sampling trajectory, we adopt the stop gradient~\cite{wu2024deep} operation to ensure the gradient flow back to the previous sampling step (we present the basic introduction in Appendix~\ref{app:background}).
The key insight is that by evaluating and optimizing the model's decisions throughout the sampling trajectory, we can learn more nuanced control over the editing process and achieve better alignment.

To cover both early and later sampling steps, we randomly select a starting timestep $r \in [0, T-k]$ and perform $k$ consecutive denoising steps, enabling gradient flow through the sampling trajectory. For each denoising step from $t$ to $t-1$, the stochastic sampling process is:
\begin{equation}
    x_{t-1} = x_t - 1/{T}s_\theta (sg(x_t), t) + \sigma_t/{\sqrt{T}} \epsilon
\end{equation}
where $\epsilon \sim \mathcal{N}(0, I)$ is standard Gaussian noise, $v_\theta(x_t, t)$ is the predicted velocity at timestep $t$, and $\sigma_t$ controls the noise injection level, $sg(.)$ denotes the stop gradient operation.
Starting from $x_r$ at timestep $r$, we sequentially apply $k$ denoising steps to obtain $x_{r-k}$,
%\begin{equation}
%    x_{r-k} = f_\theta(x_{r-1}, \ldots, f_\theta(x_r, r-1), \ldots, r-k),
%\end{equation}
each intermediate state depends on the previous one, allowing gradients to flow backward through the entire $k$-step trajectory. Calculate the log of the probability ratios across the k-step trajectory:
\begin{equation}
\small
\psi_b^{r:r-k}(\theta) = \sum_{t=r}^{r-k+1}
\log \frac{p_\theta(x_{t-1}^b \mid x_t^b, c_b)}
{p_{\theta_{\text{old}}}(x_{t-1}^b \mid x_t^b, c_b)}
\end{equation}
then the $\tilde r_b^{r:r-k}(\theta) = \exp(\mathrm{clip}(\psi_b^{r:r-k}(\theta), -\log(1+\epsilon), \log(1+\epsilon)))$.  The Dense GRPO-based optimization objective for image editing becomes:
\begin{equation}
\small
\begin{split}
    \mathcal{J}_{\text{Dense}}(\theta) = \mathbb{E}_{\tilde r, \{x_r^{b,i}\}} \left[ \frac{1}{B} \sum_{b=1}^B \min \left( \tilde r_b^{r:r-k}(\theta) \hat{A}^{b}, \,  \right. \right. \\
     \left. \left. \text{clip}(\tilde r_b^{r:r-k}(\theta), 1-\epsilon, 1+\epsilon) \hat{A}^{b} \right) - \hat{\beta} D_{\text{KL}}(\pi_\theta \parallel \pi_{\text{ref}}) \right]
\end{split}
\end{equation}
where $\hat{A}^b$ is computed using the final reward after completing the denoising from $x_{r-k}$.
\iffalse
The Dense Optimization Supervision objective becomes:
\begin{equation}
\small
\begin{split}
    \mathcal{J}_{\text{Dense}}(\theta) = \mathbb{E}_{r, \{x_r^{b}\}} \left[ \frac{1}{B} \sum_{b=1}^B \min \left( r_b^{r:r-k}(\theta) \hat{A}^{b}, \, \right. \right. \\
    \left. \left. \text{clip}(r_b^{r:r-k}(\theta), 1-\epsilon, 1+\epsilon) \hat{A}^{b} \right) - \hat{\beta} D_{\text{KL}}(\pi_\theta \parallel \pi_{\text{ref}}) \right]
\end{split}
\end{equation}
where $r_b^{r:r-k}(\theta) = \prod_{t=r}^{r-k+1} \frac{p_\theta(x_{t-1}^b | x_t^b, c_b)}{p_{\theta_{\text{old}}}(x_{t-1}^b | x_t^b, c_b)}$ is the product of probability ratios across the $k$-step trajectory, and $\hat{A}^b$ is computed using the final reward after completing the denoising from $x_{r-k}$.
\fi 
This formulation enables gradient accumulation through consecutive sampling steps, providing trajectory-level supervision rather than independent step-wise updates.
By optimizing across the full sampling trajectory rather than just the endpoint, Dense Optimization Supervision provides denser supervision signals that guide the model to make better decisions at each stage of the generation process, leading to more stable training and improved editing quality. 
\section{Experiment} \label{sec:experiment}
%In this section, we conduct comprehensive experiments to validate the effectiveness of our proposed method. We first describe our training data construction pipeline in Section~\ref{sec:training_data_construction}, followed by implementation details, evaluation metrics, and baseline methods in Section~\ref{sec:experiment_settings}. Section~\ref{sec:main_results} presents quantitative and qualitative comparisons with state-of-the-art methods, along with user studies and results on the KRIS benchmark. In Section~\ref{sec:method_analysis}, we provide in-depth analysis of our key components, including the Dense GRPO-based optimization strategy and the Dynamic Text Alignment mechanism. Finally, Section~\ref{sec:ablation_study} presents ablation studies to verify the necessity of each component in our framework.
We conduct comprehensive experiments to validate our \model's effectiveness. Section~\ref{sec:experiment_settings}  introduces the experiment settings, Section~\ref{sec:main_results} presents quantitative and qualitative comparisons, Section~\ref{sec:ablation_study} provides ablation studies and Section~\ref{sec:method_analysis} analyzes key components.

\subsection{Experimental Setup} \label{sec:experiment_settings}
\subsubsection{Training Data Construction} \label{sec:training_data_construction}

\noindent\textbf{Data Sources.} 
We construct our training dataset by combining existing editing datasets with self-constructed data, resulting in a diverse collection of image editing pairs.
Specifically, we utilize the following data sources:

\begin{itemize}
    \item \textbf{SEED-Data-Edit~\cite{ge2024seed}}~\footnote{https://huggingface.co/datasets/AILab-CVC/SEED-Data-Edit}: We sample 3k editing pairs from Part 2 of this dataset, which comprises real-world editing scenarios collected from the internet.
    \item \textbf{COCO 2017~\cite{lin2014microsoft}}~\footnote{https://cocodataset.org/}: We randomly select 1k images from the COCO 2017 dataset and construct corresponding editing pairs.
\end{itemize}
The details of data preparation process are presented in Appendix~\ref{app:data_prepare}.
%For the self-constructed COCO-based data, we employ a systematic approach to generate editing pairs. For each selected image, we apply instance segmentation to obtain object masks and select the largest segment as the target editing region based on mask area. We then create editing instructions following the template: ``Add a green segmentation mask (for object detection) to the \textit{[object name]} on the \textit{[position]} of the image'', where \textit{[object name]} and \textit{[position]} are automatically determined from the segmentation results. This construction process yields paired data consisting of the original image, the edited image with the applied mask, and the corresponding text instruction.

\begin{figure}[h]
    \centering
    \includegraphics[width=0.5\textwidth]{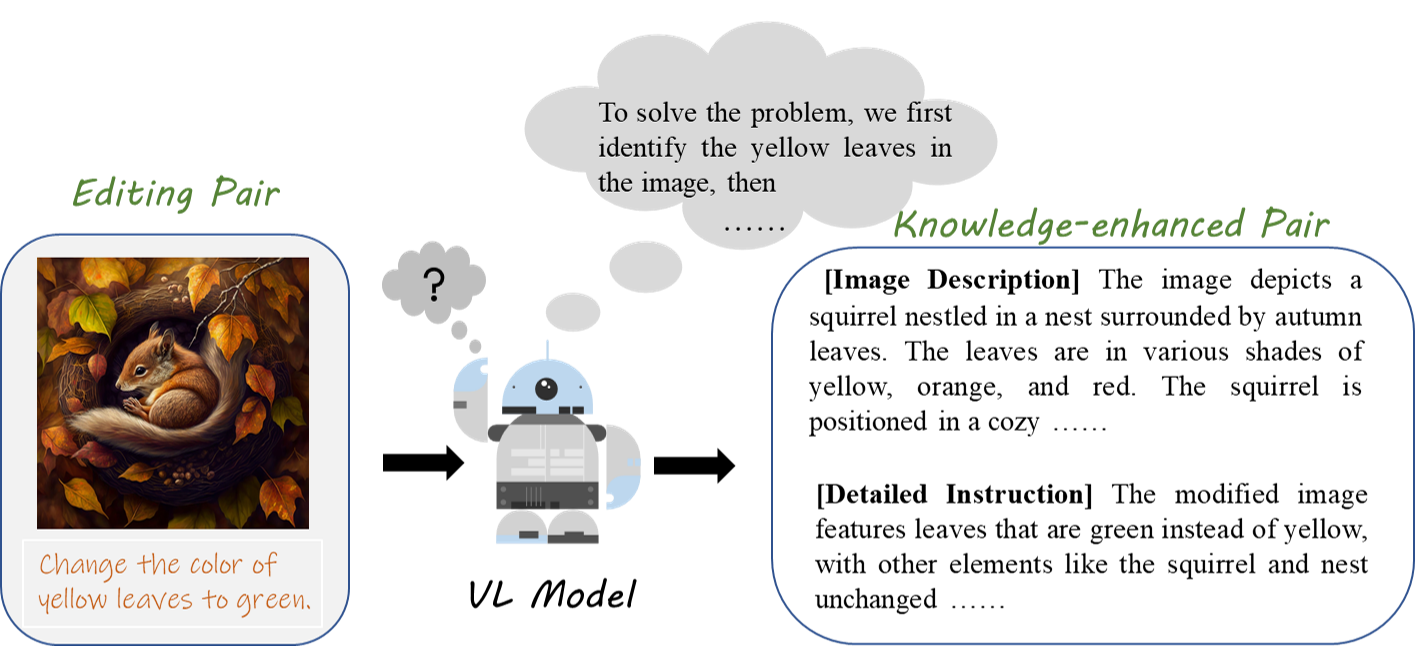}
    \caption{Illustration of the knowledge-enhanced data construction pipeline.}
    \label{fig:knowledge-enhanced-data-construction}
\end{figure}

\noindent\textbf{Knowledge-Enhanced Data Construction.} 
To improve the quality and informativeness of editing instructions, we leverage a vision-language model (VLM)~\cite{chen2025perception} to enrich the training data. 
Specifically, we feed both the source image and its corresponding editing instruction into the VLM, which generates a knowledge-enhanced instruction that incorporates additional semantic and contextual information about the scene.
This enhanced instruction replaces the original one in the training data, providing more detailed guidance for the editing model. 
Figure~\ref{fig:knowledge-enhanced-data-construction} illustrates our knowledge-enhanced data construction pipeline. The construction details and samples are presented in Appendix~\ref{app:knowledge-enhanced}.

\subsubsection{Evaluation Protocol.} 
We conduct a comprehensive evaluation from two complementary perspectives: benchmark evaluation and human assessment. The details of human assessment can be found in Appendix~\ref{app:user_study}.

\noindent\textbf{Benchmark Evaluation.}
We evaluate our \model on two complementary benchmarks that assess different aspects of image editing capabilities.
KRIS-Bench~\cite{wu2025kris}, a diagnostic benchmark designed from a cognitively informed perspective, categorizes editing tasks into three foundational knowledge types based on educational theory: factual knowledge for fact-based edits, conceptual knowledge for abstract reasoning, and procedural knowledge for multi-step operations.
The benchmark utilizes GPT-4-based assessment to measure performance across these cognitive dimensions.
Additionally, we evaluate on GEdit-Bench~\cite{liu2025step1x}, which collects authentic editing instances from real-world user scenarios to assess how well models handle genuine practical editing instructions and generalize to diverse editing challenges.  We further conduct user study to evaluate the perceptual quality and editing fidelity of our \model compared to baseline approaches in Appendix~\ref{app:user_study}.

\iffalse
\noindent\textbf{User Study.}
To assess perceptual quality and editing fidelity from a human perspective, we conduct a user study.
We randomly sample 100 editing pairs from our self-constructed test set and recruit 10 participants to evaluate the results.
Participants rate each edited image on two criteria: editing quality and instruction adherence using a 5-point Likert scale.
We then map these qualitative ratings to numerical scores.
For each editing task, we compute the mean preference score across all participants, and the overall performance is summarized by averaging scores over all editing tasks. The results can refer to Appendix~\ref{app:user_study}.
\fi 
% TODO: If using additional automatic metrics (CLIP Score, LPIPS, FID, etc.), add them here
\subsubsection{Experiment Settings.} 

\noindent\textbf{Implementation.} 
We adopt Qwen-Image-Edit~\cite{wu2025qwen}  as our base model, which provides strong vision-language understanding capabilities for image editing tasks.
To enable efficient training while maintaining model performance, we employ parameter-efficient fine-tuning (PEFT) with Low-Rank Adaptation (LoRA)~\cite{hu2022lora}.
The LoRA rank is set to 4, which strikes an effective balance between model capacity and computational efficiency.
We train our model for 500 steps using the AdamW optimizer with a learning rate of 1e-5 and a batch size of 4.
%We apply gradient accumulation over 4 steps to effectively achieve a larger batch size for more stable training.
A linear warmup schedule is employed for the first 10\% of training steps, followed by cosine learning rate decay.
To prevent overfitting, we apply gradient clipping with a maximum norm of 1.0.
The hyperparameter $k$ is set to 5 and $\xi$ is set to 64.
All experiments are conducted on 8 NVIDIA A800 GPUs with 80GB memory each, more details of the implementations can be found in Appendix~\ref{app:experiment}. %and training typically converges within 72 hours.

\noindent\textbf{Baseline Methods.} 
We compare our method against several state-of-the-art image editing approaches to demonstrate its effectiveness, including InstructPix2Pix~\cite{brooks2023instructpix2pix}, MagicBrush~\cite{zhang2023magicbrush},AnyEdit~\cite{yu2025anyedit},Step1X-Edit~\cite{liu2025step1x}, OmniGen2~\cite{wu2025omnigen2}, Qwen-Image-Edit~\cite{wu2025qwen}. Additionally, to better assess the effectiveness of our method, we equip the Qwen-Image-Edit with the VLM model to reason about the editing instructions as well. We denote this method as Qwen-Image-Edit-r1.
\iffalse
\textbf{InstructPix2Pix}~\cite{brooks2023instructpix2pix} pioneered instruction-based editing by training a conditional diffusion model on synthetically generated editing triplets.
\textbf{MagicBrush}~\cite{zhang2023magicbrush} introduces a high-quality human-annotated dataset and trains models specifically for fine-grained local editing with precise instruction following.
\textbf{AnyEdit}~\cite{yu2025anyedit} proposes a unified framework that handles diverse editing operations through a single model architecture.
\textbf{OmniGen2}~\cite{wu2025omnigen2} extends multimodal generation capabilities to editing tasks, leveraging large-scale pretraining for improved generalization.
\textbf{Qwen-Image}~\cite{wu2025qwen} serves as our base model and represents the current state-of-the-art in vision-language models with strong editing capabilities.
\fi 
We compare against these baselines using their official implementations and pretrained checkpoints to ensure fair evaluation.

% TODO: List baseline methods with citations

\subsection{Main Results} \label{sec:main_results}

\begin{table*}[h]
    \centering
     \caption{Results on Kris-Bench, where VC is the visual consistency score, VQ is the visual quality score, IF is the instruction following score, KP is the knowledge preservation score, and Avg is the average score.}
    \resizebox{\linewidth}{!}{
    \begin{tabular}{c|ccccc|ccccc|ccccc}
        \toprule
        \multirow{2}{*}{\textbf{Method}} & \multicolumn{5}{c|}{\textbf{Social Science Science}} &\multicolumn{5}{c|}{\textbf{Natural Science}} &  \multicolumn{5}{c}{\textbf{Knowledge Reasoning}} \\
       
        ~& VC & VQ & IF & KP & Avg & VC & VQ & IF & KP & Avg & VC & VQ & IF & KP & Avg  \\
        \midrule
        \textbf{InsPix2Pix} & 15.75 & 50.00 & 14.25 & 10.25 & 22.56 & 18.75 & 58.25 & 17.50 & 11.75 & 26.56 & 31.00 & 84.00 & 5.50 & 3.50 & 31.00\\
        %\midrule
        \textbf{MagicBrush} & 54.00 & 70.00 & 27.25 & 20.50 & 42.94 & 47.00 & 72.25 & 19.00 & 13.50 & 38.06 & 62.50 & 88.75 & 11.25 & 8.25& 42.69\\
        %\midrule
        \textbf{AnyEdit} & 62.00 & 66.75 & 15.00 & 10.50 & 38.36 & 61.75 & 77.75 & 18.25 & 14.00 & 42.94 & 68.25 & 84.50 & 12.00 & 11.00 & 43.94 \\
        %\midrule
        \textbf{Step1X-Edit} & 78.50 & 77.75 & 28.75 & 22.75 & 51.94 & 82.75 & 79.00 & 28.00 & 21.00 & 52.69 & 78.75 & 81.25 & 20.00 & 14.00 & 48.50\\
        \textbf{OmniGen2} & 75.40&87.60&20.45 & 16.32 & 50.46 & 65.59 & 88.11 & 22.34 & 14.03 & 47.76 & 39.67 & 78.90 & 11.90 & 5.17 & 33.90 \\
        \textbf{Qwen-Image-Edit} & 72.20 & 86.00 & 55.94 & 50.82 & 66.40 & 67.05 & 83.52 & 44.08 & 33.72 & 57.30 & 74.00 & 85.67 & 29.66 & 22.41 & 53.39\\
        \midrule
        \textbf{Qwen-Image-Edit-r1} &  79.20 & 89.14 & \textbf{73.17} & \textbf{70.53} & \textbf{77.99} &  \textbf{72.07} & 87.60 & 52.88 & 46.40 & 64.78 & 71.33 & 88.37 & 35.03 & 27.55 & 55.48\\
        \textbf{\model} & \textbf{80.60} & \textbf{92.40} & 70.25 & 66.74 & 77.64 & 70.36 & \textbf{90.47 }&\textbf{57.59} & \textbf{50.92} & \textbf{67.42} &\textbf{ 71.50} &\textbf{ 91.72 }& \textbf{40.14 }& \textbf{33.33 }& \textbf{59.29 } \\
        \bottomrule
    \end{tabular}
   }
    \label{tab:kris_bench_results}
\end{table*}
\subsubsection{Quantitative Comparison.}
% TODO: Add quantitative results comparing with baselines
%\textbf{Results on Kris-Bench.} 
Table~\ref{tab:kris_bench_results} presents quantitative results on Kris-Bench, four metrics: VC (visual consistency), VQ (visual quality), IF (instruction following), and KP (knowledge preservation) averaged over all three domains (social science, natural science, and knowledge reasoning), evaluated by GPT-4.. Our \model achieves the highest visual quality scores across all three domains (social science, natural science, and knowledge reasoning), demonstrating superior perceptual quality while maintaining competitive performance in other metrics. In social science scenarios, we rank among the top-performing approaches, trailing only Qwen-Image-Edit-r1 in overall score while maintaining the lead in visual quality, and show strong performance across natural science and knowledge reasoning domains, validating effectiveness on diverse editing tasks requiring domain knowledge and reasoning capabilities. These results demonstrate that our dense optimization strategy effectively balances visual quality with instruction following, addressing the fundamental challenge of generating perceptually superior edits while maintaining semantic accuracy and content fidelity. The results of GEdit-Bench can be found in Appendix~\ref{app:res_gedit}.
\subsubsection{Qualitative Comparison.}
% TODO: Add qualitative visualizations comparing with baselines
\begin{figure*}
    \centering
    \includegraphics[width=\linewidth]{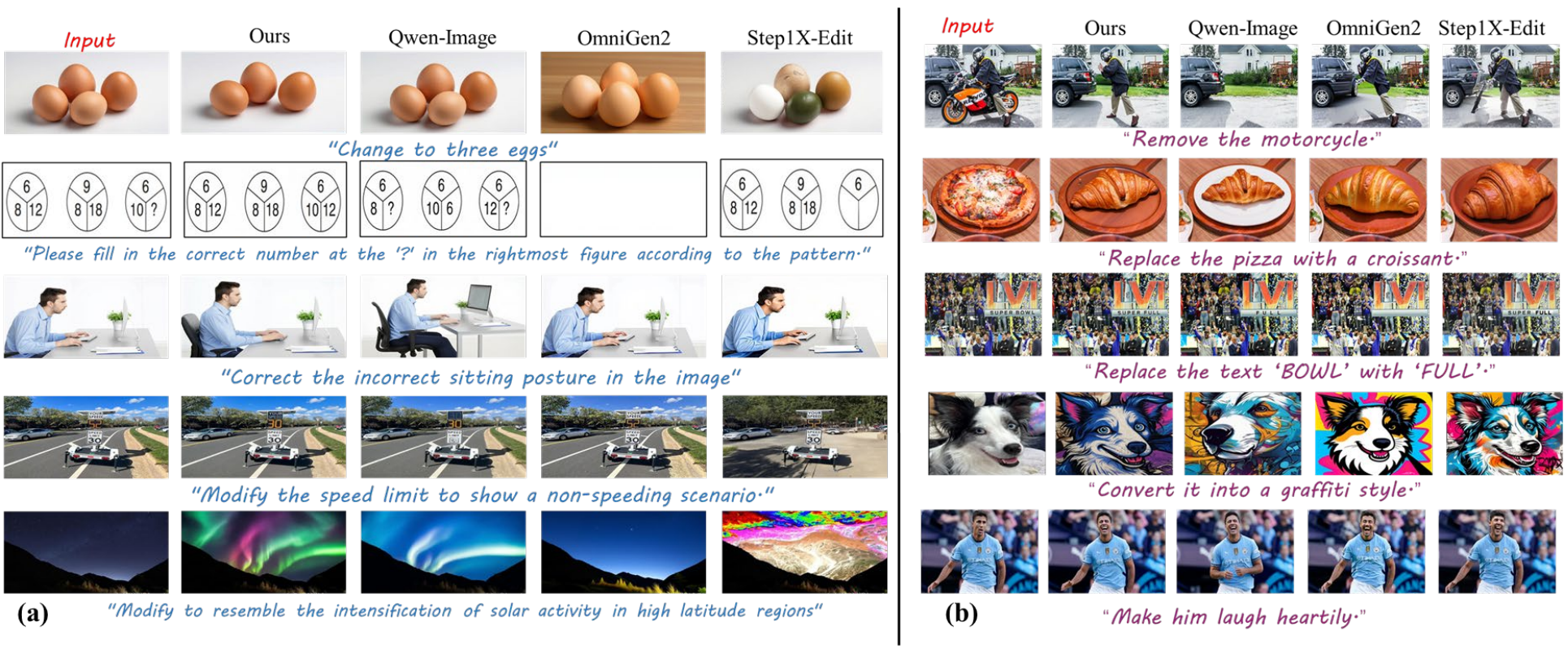}
    \caption{Qualitative comparison on Kris-Bench (a) and GEdit-Bench (b). Our \model achieves superior fine-grained instruction following while maintaining high visual quality and natural blending of edited regions.}
    \label{fig:qualitative_results}
\end{figure*}
Figure~\ref{fig:qualitative_results} presents qualitative comparisons with baseline methods. On Kris-Bench (Figure~\ref{fig:qualitative_results}(a)), our \model demonstrates superior fine-grained instruction following: when instructions specify precise colors, positions, or quantities, our method accurately captures these details while maintaining high visual quality. In contrast, baseline methods often miss specific attributes or produce visually inconsistent results. This confirms that our dense reward optimization effectively guides the model toward both semantic accuracy and perceptual quality. On GEdit-Bench (Figure~\ref{fig:qualitative_results}(b)), our \model preserves general editing capabilities across diverse scenarios from object removal to style transfer, with edited regions blending naturally with unchanged areas. This demonstrates that our specialized optimization for fine-grained alignment does not compromise general editing proficiency. 
\subsection{Ablation Study} \label{sec:ablation_study}
\begin{figure}
    \centering
    \includegraphics[width=\linewidth]{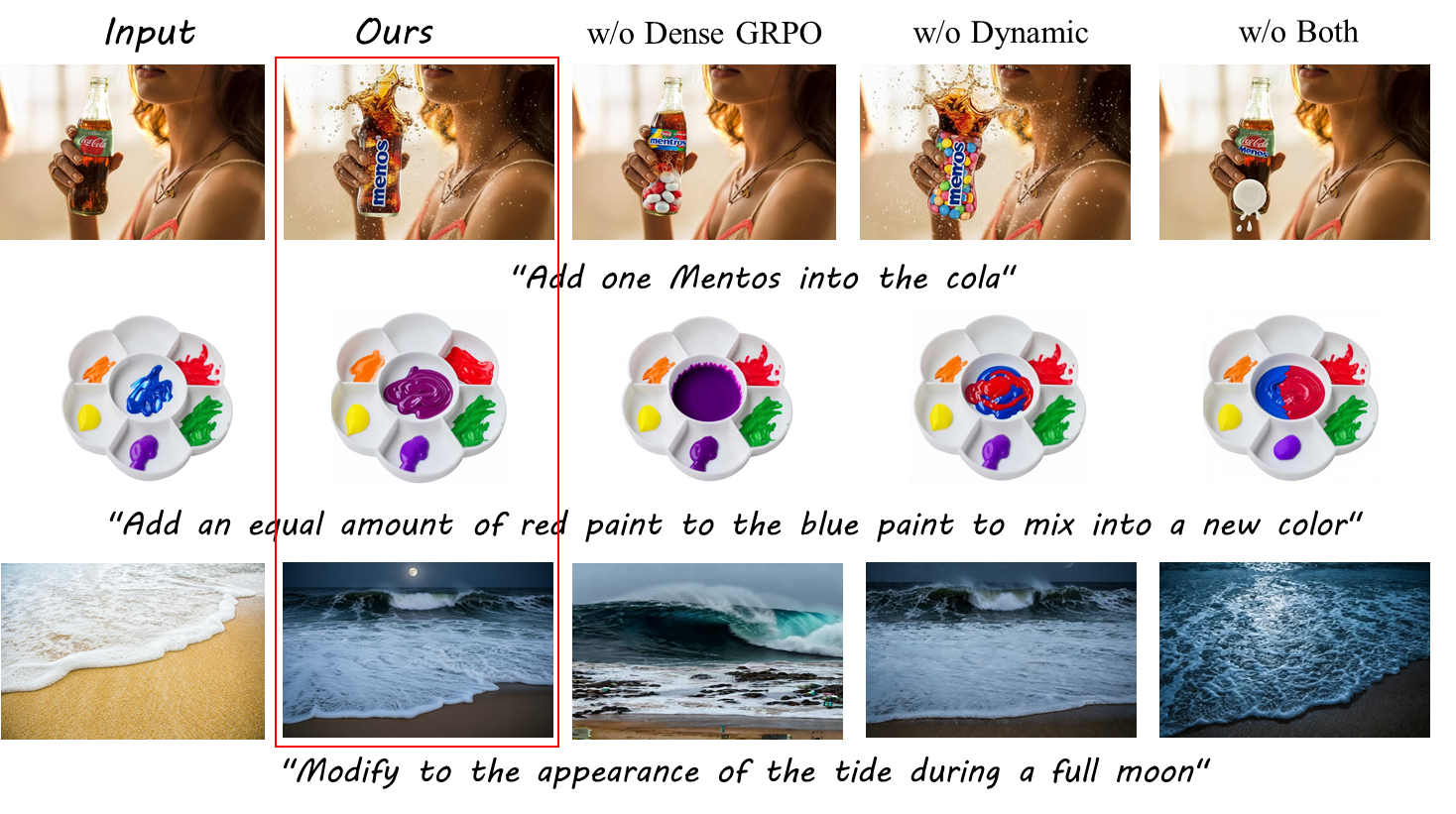}
    \caption{The visualization results of the ablation study.}
    \label{fig:ablation_study_visualization}
\end{figure}
To systematically validate the contribution of each component in our \model, we conduct comprehensive ablation studies on the Knowledge Reasoning category of the KRIS benchmark.
We evaluate three configurations: (1) removing Dense GRPO-based optimization while retaining Dynamic Text Alignment, (2) removing Dynamic token focus relocation while retaining Dense GRPO-based optimization, and (3) removing both components, which effectively reverts to the base Qwen-Image-Edit-r1 model performance.
This experimental design allows us to isolate the individual and synergistic effects of our proposed mechanisms.

\begin{table}[h]
    \centering
        \caption{Ablation study results on the Knowledge Reasoning category. Each row shows performance when specific components are removed, demonstrating their individual contributions.}
    \resizebox{\linewidth}{!}{    \begin{tabular}{c|ccccc}
        \toprule
        \textbf{Method}  & \multicolumn{5}{c}{\textbf{Knowledge Reasoning}} \\
        ~ & \textbf{VC} & \textbf{VQ} & \textbf{IF} & \textbf{KP} & \textbf{Avg} \\
        \midrule
        w/o Dense GRPO & 41.67 &	75.00 &	35.04	& 32.72 & 46.11\\
        w/o Dynamic & 62.67&	86.17	&41.73	& 36.09& 54.66\\
        w/o both (Base Model) &  71.33 & 88.37 & 35.03 & 27.55 & 55.48 \\
        \midrule 
        \textbf{\model (Full Model)} &\textbf{71.50} & \textbf{91.72} & \textbf{40.14} & \textbf{33.33} & \textbf{59.29}  \\
        \bottomrule
    \end{tabular}
    }
    \label{tab:ablation}
\end{table}
Table~\ref{tab:ablation} demonstrates that both components contribute substantially to our proposed \model's performance. Removing Dense GRPO-based optimization causes the most significant degradation, confirming its critical role in learning fine-grained semantic correspondences through trajectory-level supervision. Removing Dynamic Token Focus Relocation leads to moderate performance reduction, demonstrating its importance for hierarchical semantic understanding. The full model achieves 59.29, representing a notable improvement over the base model, which validates the synergistic effect of both mechanisms. Notably, the base model (w/o both) outperforms ``w/o Dynamic'' alone, indicating that Dynamic Token Focus Relocation is most effective when combined with Dense GRPO's trajectory-level optimization, this complementary relationship highlights the importance of jointly optimizing attention patterns and reward signals for fine-grained editing.

\subsection{Analysis of Key Components}  \label{sec:method_analysis}

\subsubsection{Dense GRPO-based optimization.}  \label{sec:analy_dense}
\begin{figure}[h]
    \centering
    \begin{subfigure}{0.23\textwidth}
         \includegraphics[width=\linewidth]{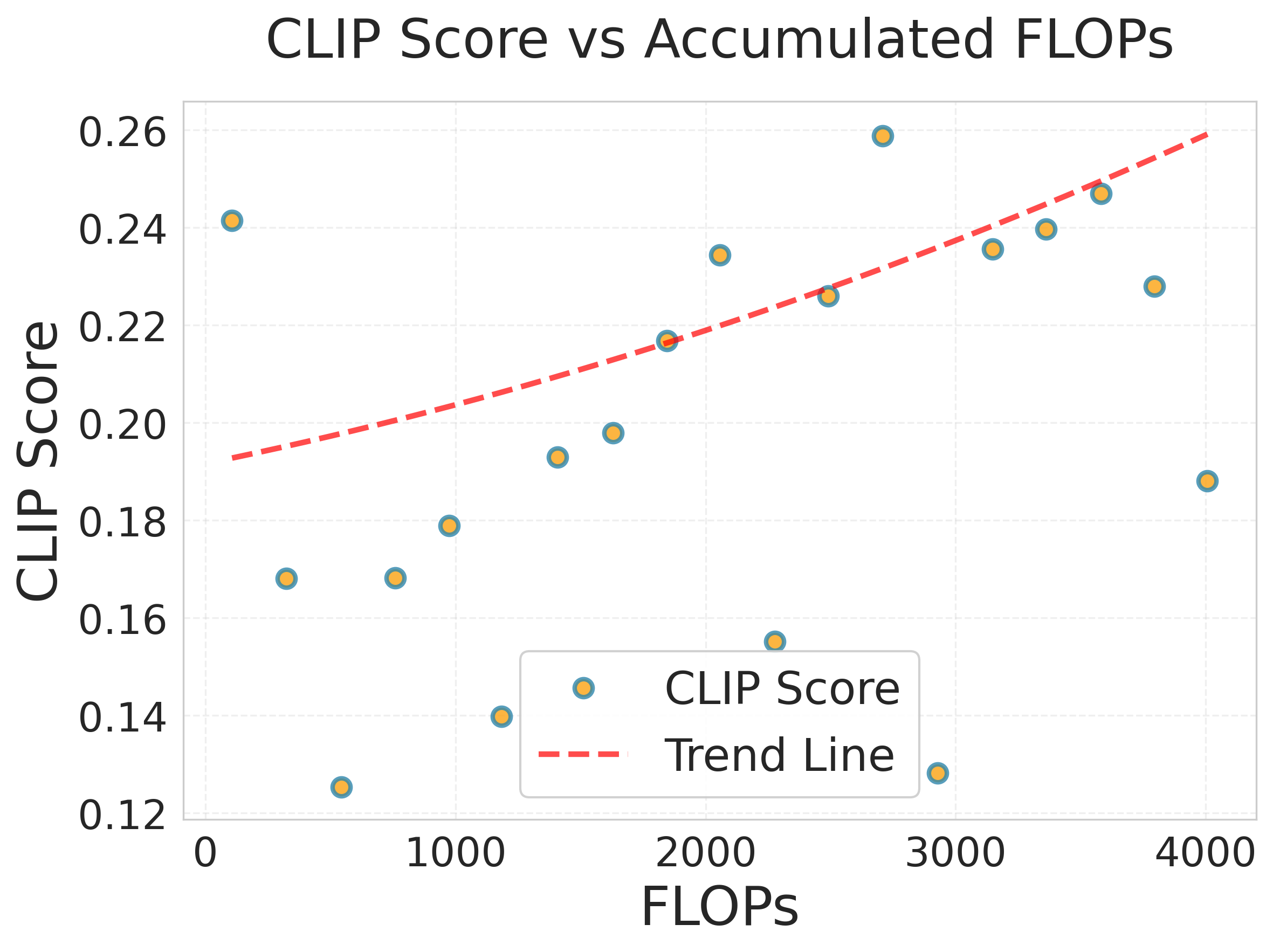}
    \caption{The CLIP score vs. FLOPs when with Dense GRPO.}
     \label{fig:dense_grpo_analysis_clip}
    \end{subfigure}
    \hspace{2pt}
    \begin{subfigure}{0.23\textwidth}
         \includegraphics[width=\linewidth]{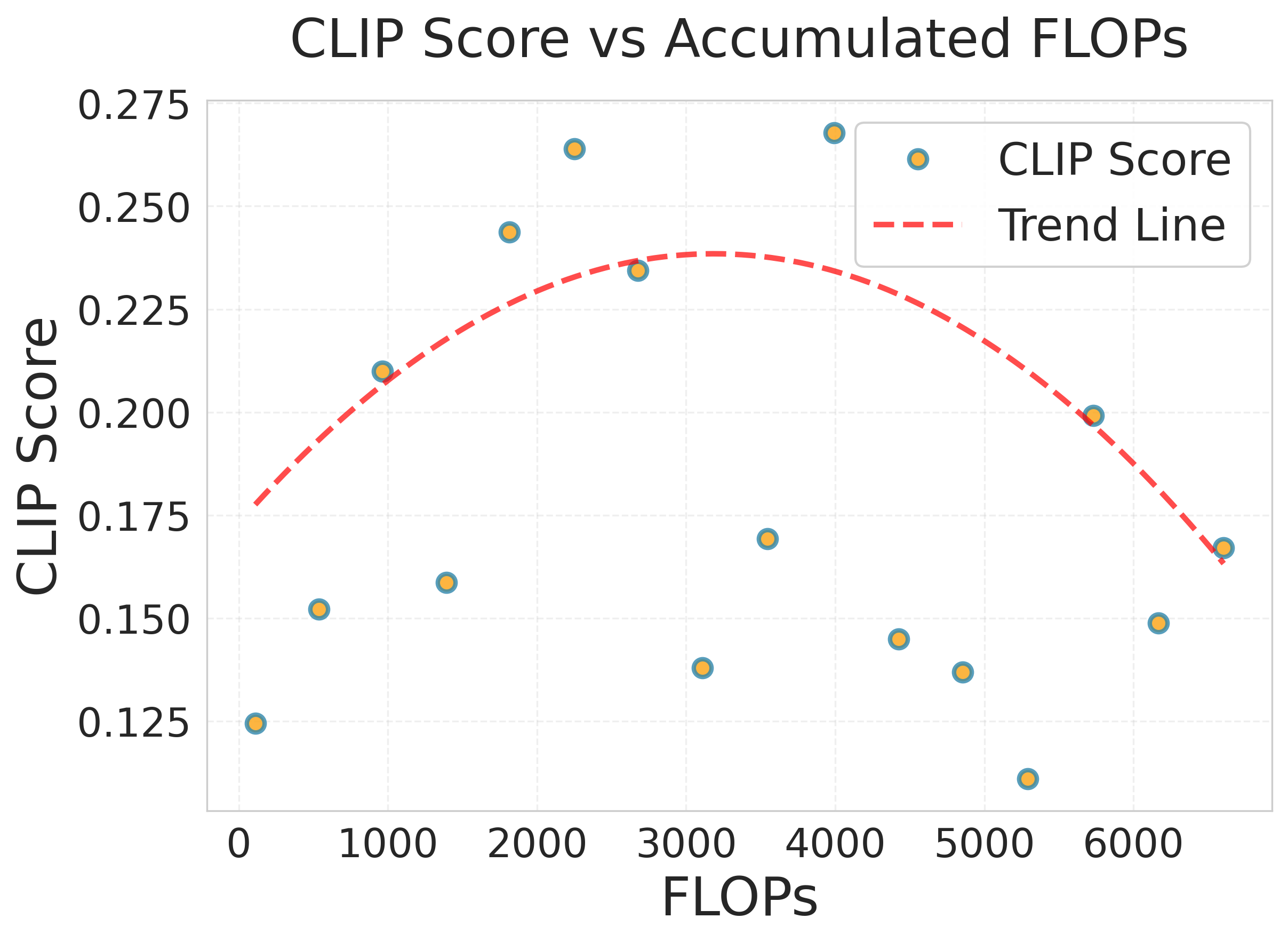}
    \caption{The CLIP score vs. FLOPs when with standard GRPO.}
    \label{fig:standard_grpo_analysis_clip}
    \end{subfigure}
   %\label{fig:dense_grpo_analysis}
   \vspace{-12pt}
\end{figure}
To demonstrate the effectiveness of Dense GRPO, we analyze computational efficiency and optimization performance compared to standard GRPO. As shown in Figures~\ref{fig:dense_grpo_analysis_clip} and~\ref{fig:standard_grpo_analysis_clip}, Dense GRPO achieves superior performance with reduced computational overhead, exhibiting a steeper learning curve and greater training stability. This improvement stems from dense gradient flow through intermediate sampling steps, which provides more informative supervision signals throughout the generation trajectory. In contrast, standard GRPO shows erratic CLIP score fluctuations and limited capacity for fine-grained semantic correspondences due to sparse gradient signals computed only at sampling endpoints. By propagating gradients through the entire sampling process, Dense GRPO enables more effective optimization dynamics and better instruction-image alignment.

\subsubsection{Dynamic Text Alignment Mechanism.} \label{sec:analy_dynamic}

\begin{figure}[h]
    \centering
    \includegraphics[width=0.9\linewidth]{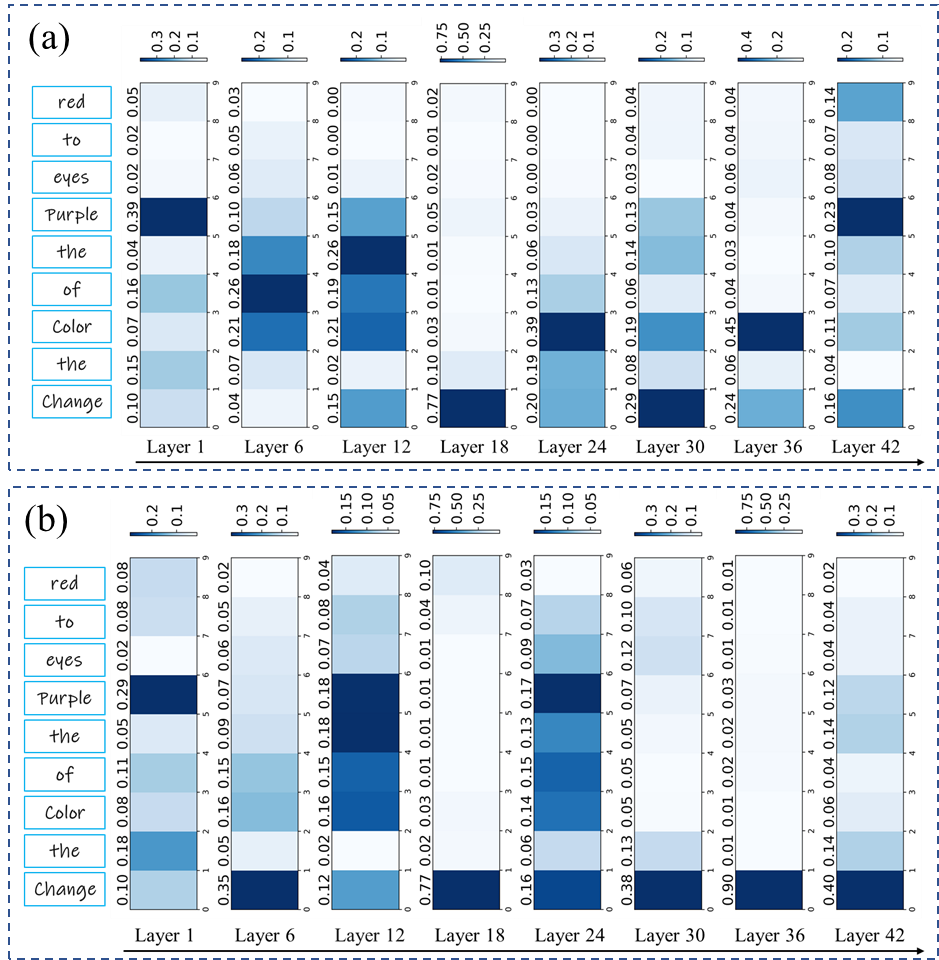}
    \caption{The attention maps from intermediate layers and the corresponding text importance scores with our proposed dynamic text alignment mechanism.}
    \label{fig:text_alignment_analysis}
    \vspace{-8pt}
\end{figure}
To analyze the effectiveness of Dynamic Token Focus Relocation, we conduct layer-wise analysis of attention patterns. Figure~\ref{fig:text_alignment_analysis} compares attention distributions with (a) and without (b) our mechanism. With dynamic relocation, the model exhibits hierarchical attention allocation: for the instruction ``Change the color of the purple eyes to red,'' early layers prioritize abstract concepts (``the color''), intermediate layers shift to specific attributes (``purple,'' ``red''), and deeper layers concentrate on actions (``change''). This hierarchical progression enables compositional understanding and precise semantic grounding. In contrast, without dynamic relocation, attention persistently focuses on ``change'' throughout all layers, resulting in incomplete semantic parsing and limited editing capabilities. These results demonstrate that dynamic attention reweighting across network depth effectively prioritizes semantically salient tokens at appropriate stages, yielding more accurate instruction-faithful editing.

\section{Conclusion} \label{sec:conclusion}
We present a unified framework \model, for fine-grained instruction-based image editing that addresses two fundamental limitations: semantic understanding of complex instructions and sparse optimization feedback. Our approach combines multi-modal reasoning with dense reward optimization through three key components. First, we leverage Multi-modal Large Language Models to decompose complex instructions into semantically correct editing directives. Second, we introduce Dynamic Token Focus Relocation to enable hierarchical attention to fine-grained attributes across network layers. Third, we develop Dense GRPO-based optimization that propagates gradients across consecutive denoising steps, enabling trajectory-level supervision through the sampling process. Extensive experiments on Kris-Bench and GEdit-Bench demonstrate that our method achieves state-of-the-art performance in fine-grained instruction following while preserving general editing capabilities. These results validate that dense supervision across the sampling trajectory effectively balances precise attribute control with visual quality, providing a principled approach for instruction-based image editing.
{
    \small
    \bibliographystyle{ieeenat_fullname}
    \bibliography{main}
}

\clearpage
\setcounter{page}{1}
\maketitlesupplementary

\section{Algorithm}
We provide the training procedure for \model, which combines Dynamic Token Focus Relocation with Dense GRPO Optimization.

\begin{algorithm}[h]
\caption{Training Procedure of \model}
\label{alg:intelliedit}
\begin{algorithmic}[1]
\REQUIRE Policy $\pi_\theta$, reference $\pi_{\text{ref}}$, reward model $R$, MLLM, batch size $B$, group size $G$, timesteps $T$, dense steps $k$
\STATE Initialize $\theta$, token focus predictor $p_\eta$, soft tokens $\{s_i^{1:k}\}_{i=1}^N$
\FOR{each training iteration}
    \STATE Sample batch $\{(\text{image}_b, \text{instruction}_b)\}_{b=1}^B$
    \STATE Decompose instructions: $c_b \leftarrow$ MLLM.decompose($\text{instruction}_b$) for $b=1,\ldots,B$
    \STATE \textbf{Dynamic Token Focus:} For each layer $i$, predict $\text{pos} \leftarrow p_\eta(h_i^l)$ and inject $h_i^{\text{pos}:\text{pos}+k} \leftarrow s_i^{1:k}$
    \FOR{$b = 1$ to $B$}
        \STATE Sample $r \sim \text{Uniform}[k, T]$ and initialize $x_r^{b,1:G}$
        \FOR{$i = 1$ to $G$}
            \STATE Denoise $k$ steps: $x_{t-1}^{b,i} \leftarrow x_t^{b,i} - \frac{1}{T}(v_\theta(\text{sg}(x_t^{b,i}), t) + \frac{\sigma_t^2}{2}[x_t^{b,i} + (1-t)v_\theta(\text{sg}(x_t^{b,i}), t)]) + \frac{\sigma_t}{\sqrt{T}}\epsilon$ for $t=r,\ldots,r-k$
            \STATE Complete: $x_0^{b,i} \leftarrow \text{sg}(\text{Denoise}(x_{r-k}^{b,i}))$
            \STATE Evaluate: $R(x_0^{b,i}, c_b)$
        \ENDFOR
    \ENDFOR
    \STATE Compute the batch advantages: $\hat{A}^b \leftarrow \frac{1}{G}\sum_{i=1}^G \frac{R(x_0^{b,i}, c_b) - \mu_{\text{batch}}}{\sigma_{\text{batch}}}$ where $\mu_{\text{batch}}, \sigma_{\text{batch}}$ are batch statistics
    \STATE Compute probability ratios: $\tilde{r}_b^{r:r-k}(\theta) \leftarrow \exp(\text{clip}(\sum_{t=r}^{r-k+1} \log \frac{p_\theta(x_{t-1}^b | x_t^b, c_b)}{p_{\theta_{\text{old}}}(x_{t-1}^b | x_t^b, c_b)}, -\log(1+\epsilon), \log(1+\epsilon)))$
    \STATE Compute optimization objective: $\mathcal{J}_{\text{Dense}}(\theta) \leftarrow \frac{1}{B}\sum_{b=1}^B \min(\tilde{r}_b^{r:r-k}\hat{A}^b, \text{clip}(\tilde{r}_b^{r:r-k}, 1-\epsilon, 1+\epsilon)\hat{A}^b) - \hat{\beta} D_{\text{KL}}(\pi_\theta \parallel \pi_{\text{ref}})$
    \STATE Update: $\theta, \eta, \{s_i^{1:k}\}_{i=1}^N \leftarrow$ gradient step on maximizing $\mathcal{J}_{\text{Dense}}(\theta)$
\ENDFOR
\RETURN $\pi_\theta$, $p_\eta$, $\{s_i^{1:k}\}_{i=1}^N$
\end{algorithmic}
\end{algorithm}

The algorithm integrates three key components: (1) MLLM-based instruction decomposition for semantic understanding, (2) Dynamic Token Focus Relocation for hierarchical attention patterns, and (3) Dense GRPO Optimization with gradient accumulation across $k$ consecutive denoising steps. The batch-level advantage computation stabilizes training for editing tasks, while stop gradient operations ensure proper gradient flow through the sampling trajectory.

\section{Background} \label{app:background}
\subsection{Stop Gradient}
Stop gradient is a technique that selectively blocks gradient flow through specific computational paths during backpropagation.
In neural network training, gradients typically flow backward through all operations to update model parameters.
However, in certain scenarios, allowing unrestricted gradient flow can lead to training instability or undesired parameter updates.
By applying stop gradient operations, denoted as $\text{sg}(\cdot)$, we can treat certain values as constants during the backward pass while preserving their computational graph during the forward pass.

Formally, for a tensor $x$, the stop gradient operation is defined as:
\begin{equation}
    y = \text{sg}(f(x)), \quad \text{where} \quad \frac{\partial y}{\partial x} = 0
\end{equation}
During forward propagation, $y = f(x)$, but during backpropagation, no gradients flow through this operation.
This is particularly useful in reinforcement learning and self-supervised learning scenarios where we want to compute values based on the current model state without updating certain components.

In our framework, we strategically employ stop gradient operations to stabilize training when computing target values or baseline estimates that should not directly influence certain model components through gradient descent, as we detail in Section~\ref{sec:dense_grpo}.
\begin{figure*}[htbp]
    \centering
    \begin{subfigure}{\linewidth}
        \includegraphics[width=0.95\linewidth]{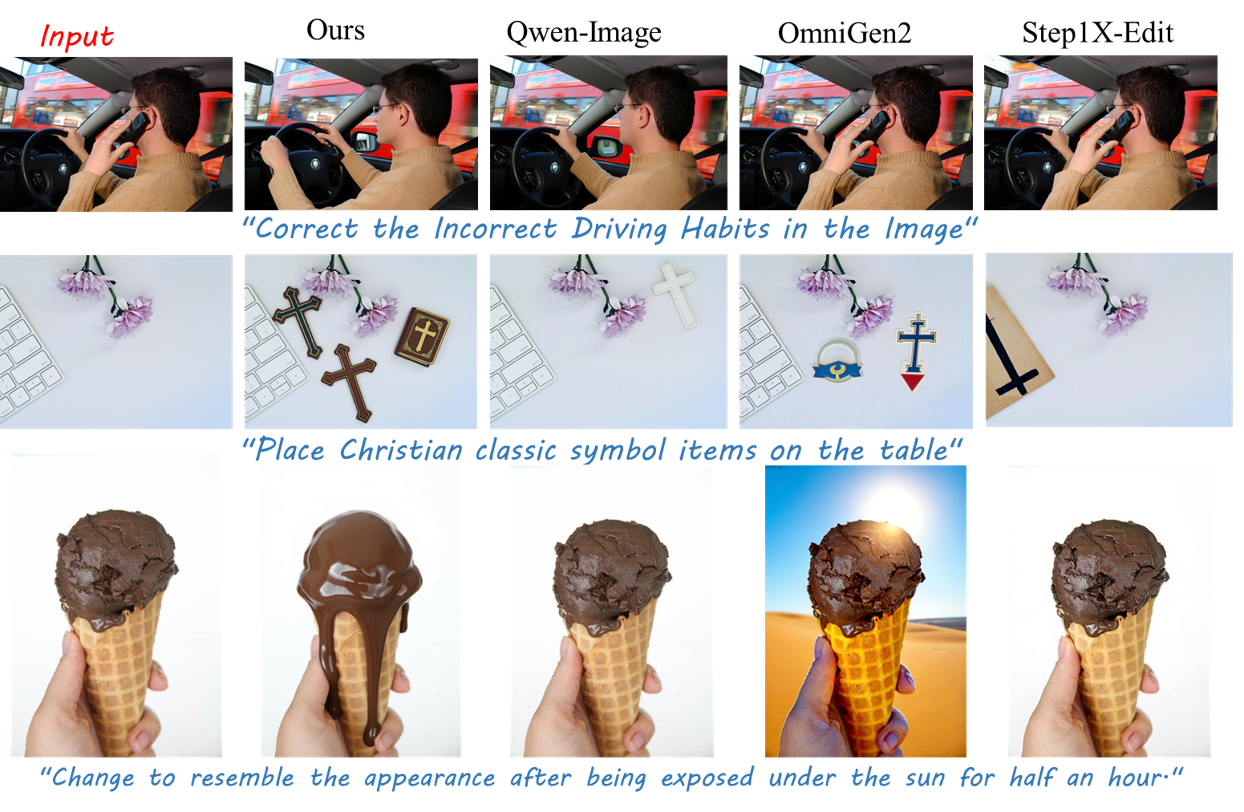}
    \end{subfigure}
    \begin{subfigure}{\linewidth}
        \includegraphics[width=0.95\linewidth]{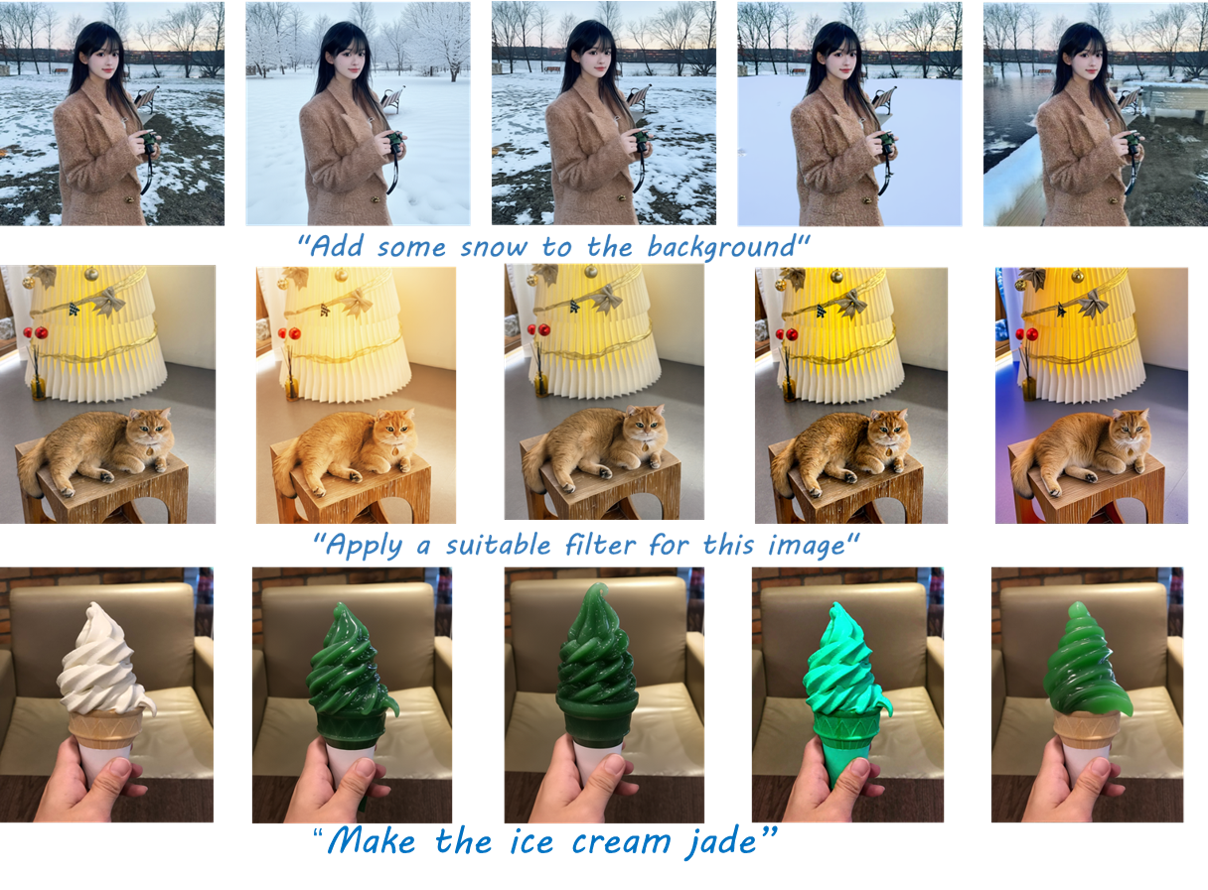}
    \end{subfigure}
\end{figure*}

\begin{figure*}[htbp]
    \centering
      \begin{subfigure}{\linewidth}
        \includegraphics[width=\linewidth]{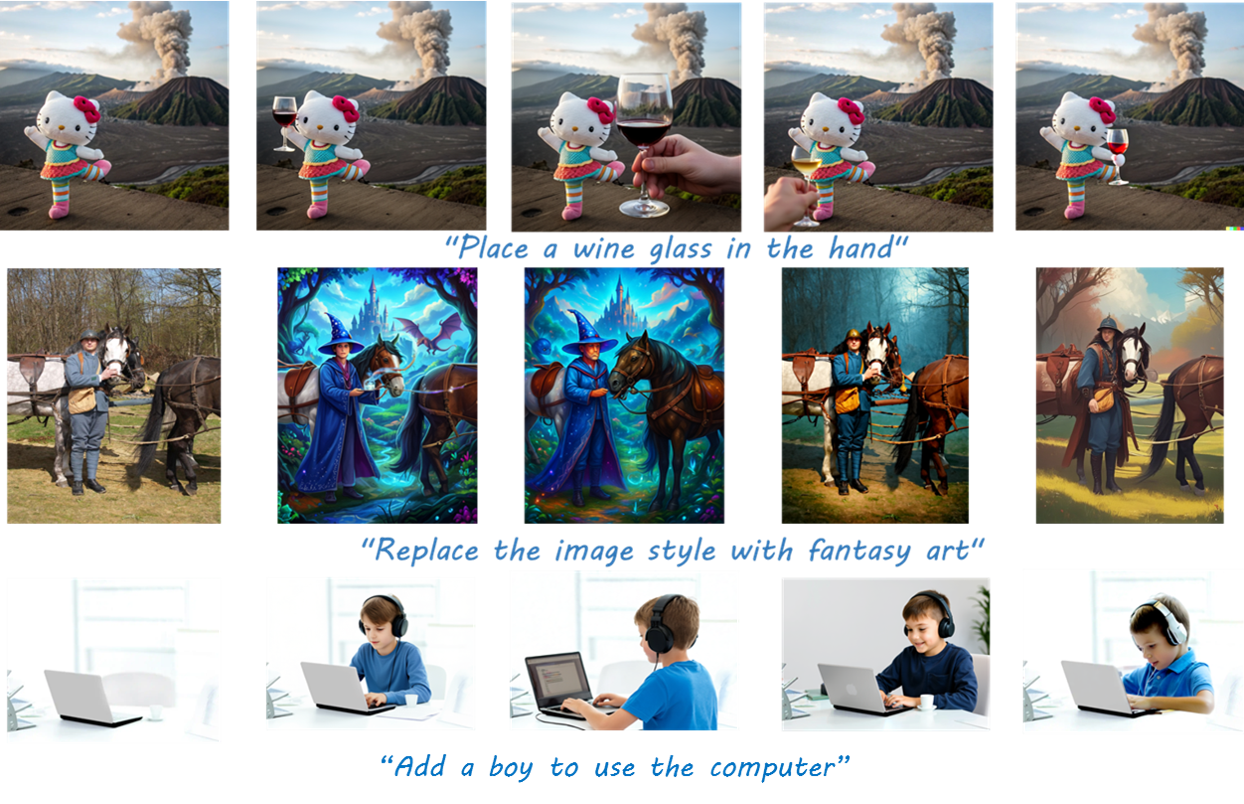}
    \end{subfigure}
    \caption{Qualitative comparison with instruction-based editing methods. \model maintains high visual quality and natural blending between edited and unedited regions, while baseline methods struggle with consistency preservation.}
    \label{fig:comparison_instruction_based}
\end{figure*}
\section{Methodology} \label{app:method}

\subsection{Visual Reasoning Enhanced Text Encoding} \label{app:reasoning_text}
Visual reasoning enhanced text encoding leverages vision-language models (VLMs) to generate enriched text representations that exhibit superior alignment with visual content.
Recent research has demonstrated that exploiting the reasoning capabilities of VLMs to extract detailed, fine-grained visual information yields more effective guidance for model behavior~\cite{yeh2025beyond,wang2024genartist}.
This approach addresses the limitation of naive text encoding, which often fails to capture the nuanced semantic relationships between instructions and visual content necessary for precise editing operations.

The encoding process follows a structured reasoning pipeline.
Given an editing instruction and source image, the VLM performs hierarchical semantic analysis through three stages: (1) \textbf{Visual Comprehension} -- generating a detailed image description encompassing objects, quantities, textual elements, spatial relationships, and salient visual features; (2) \textbf{Reasoning} -- constructing a logical reasoning chain that connects the visual content with the editing instruction; (3) \textbf{Instruction Synthesis} -- producing an enriched editing instruction that incorporates both the original command and contextual visual information.

To further enhance editing precision, we extend this standard reasoning framework with an additional \textbf{Target Specification} stage that explicitly identifies and describes the specific image regions or objects to be modified.
This four-stage reasoning process provides comprehensive semantic grounding for the editing model, enabling more accurate interpretation of editing intents.

Formally, we prompt the VLM to generate structured outputs following the format:
\begin{itemize}
    \item \textbf{Caption} (\texttt{<info>}): A comprehensive visual description capturing objects, quantities, text, spatial relations, and salient features;
    \item \textbf{Reasoning} (\texttt{<think>}): A logical reasoning chain connecting visual content with the editing instruction;
    \item \textbf{Answer} (\texttt{<answer>}): An enriched editing instruction synthesizing original command and visual context;
    \item \textbf{Target} (\texttt{<object>}): Detailed specification of regions or objects to be modified.
\end{itemize}

The complete output structure is thus: 
\begin{equation}
\begin{aligned}
    \texttt{<info>...</info> <think>...</think>} \\
    \texttt{<answer>...</answer> <object>...</object>}
\end{aligned}
\end{equation}
,which provides the editing model with multi-level semantic guidance ranging from holistic scene understanding to precise target localization.

\begin{figure*}[htbp]
    \centering
    \begin{subfigure}{\linewidth}
        \includegraphics[width=\linewidth]{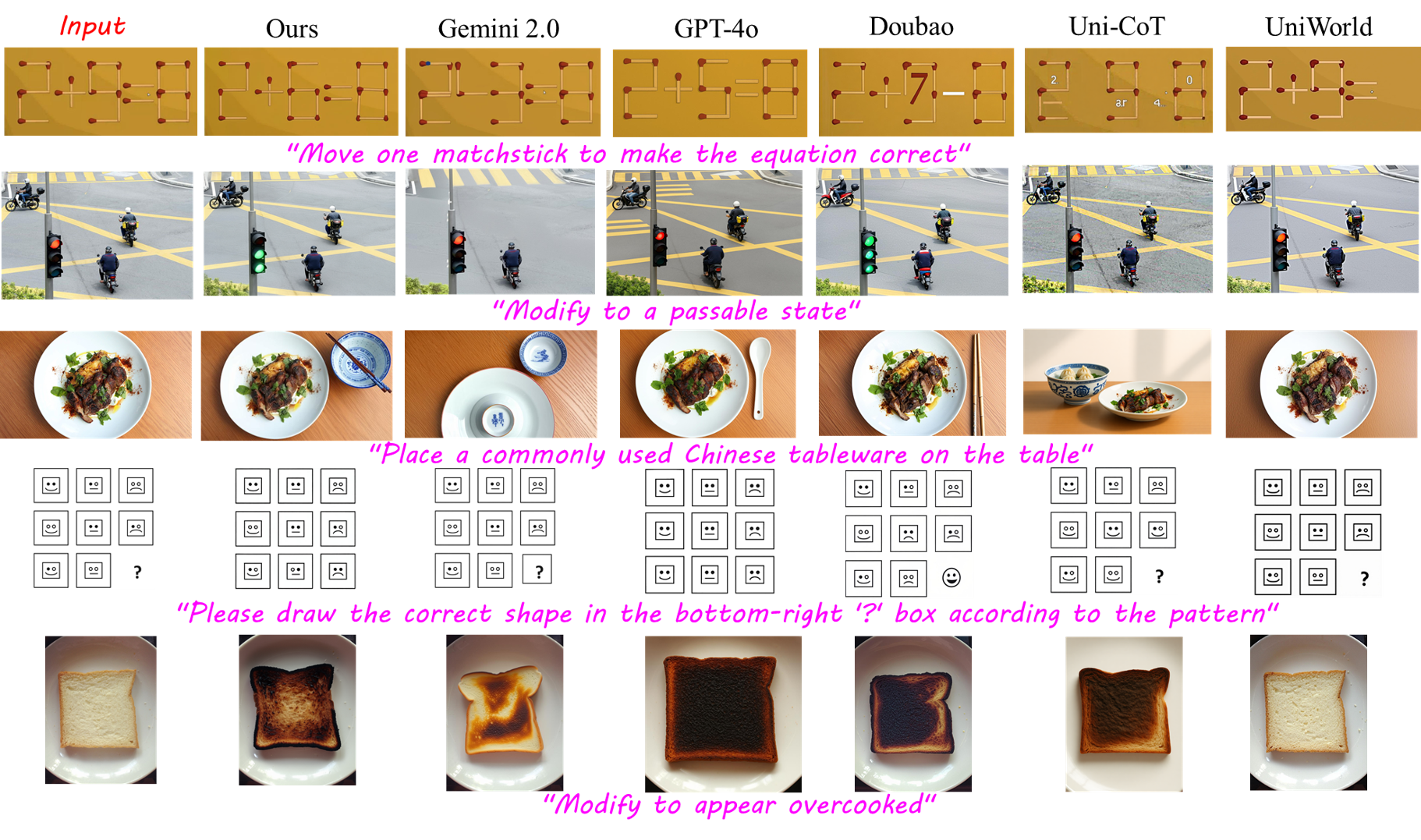}
    \end{subfigure}
    \begin{subfigure}{\linewidth}
        \includegraphics[width=\linewidth]{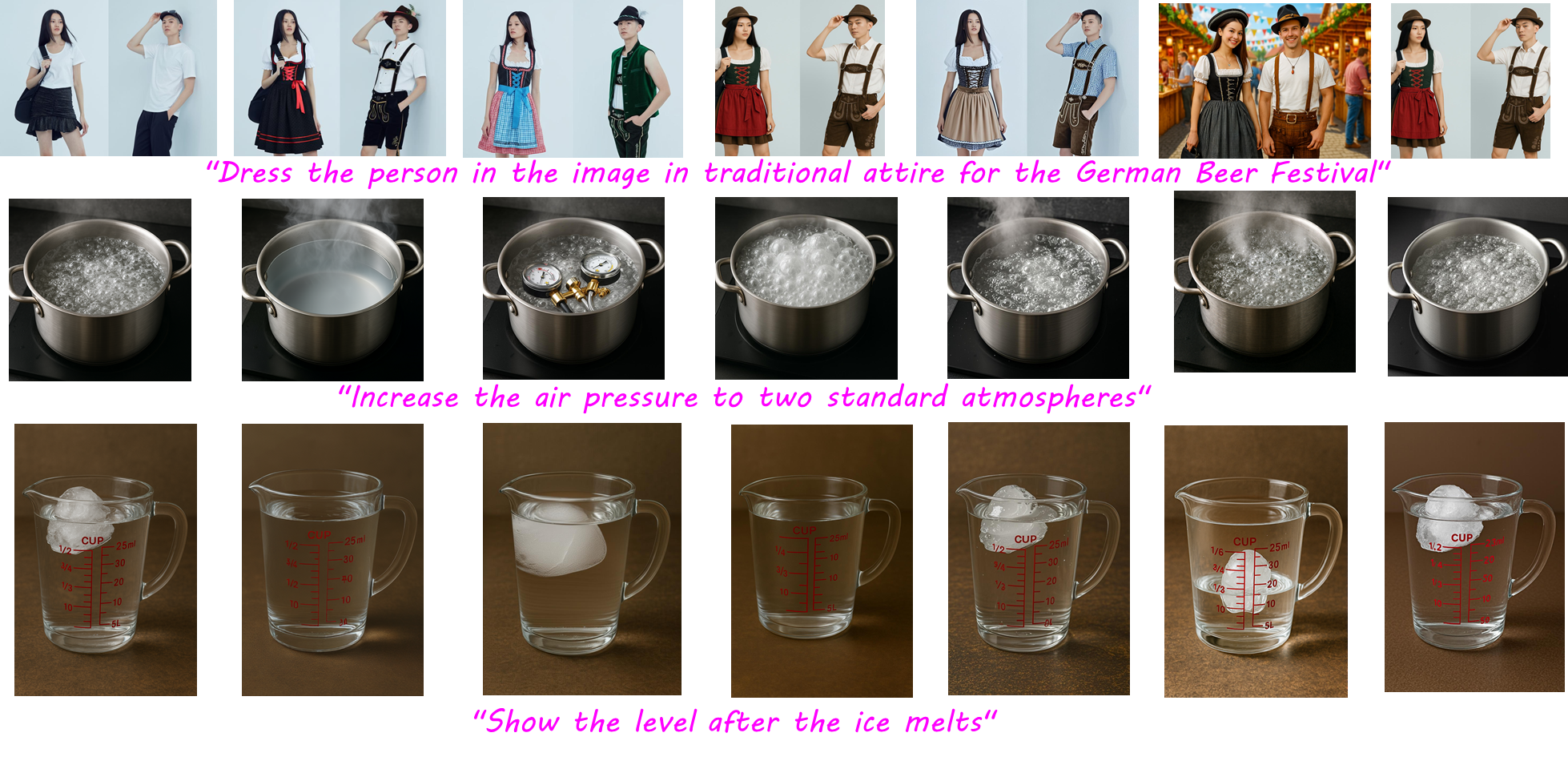}
    \end{subfigure}
    \caption{Qualitative comparison with knowledge-based editing methods. Each example shows source image, instruction requiring domain knowledge or reasoning, and results from different methods.}
    \label{fig:comparison_knowledge_based}
\end{figure*}

\subsection{Data Preparation} \label{app:data_prepare}
We construct our training dataset by combining existing editing datasets with self-constructed data, resulting in a diverse collection of image editing pairs.
Specifically, we utilize the following data sources:

\begin{itemize}
    \item \textbf{SEED-Data-Edit~\cite{ge2024seed}}~\footnote{https://huggingface.co/datasets/AILab-CVC/SEED-Data-Edit}: We sample 3k editing pairs from Part 2 of this dataset, which comprises real-world editing scenarios collected from the internet.
    \item \textbf{COCO 2017~\cite{lin2014microsoft}}~\footnote{https://cocodataset.org/}: We randomly select 1k images from the COCO 2017 dataset and construct corresponding editing pairs.
\end{itemize}

\begin{figure*}[h]
    \centering
    \begin{subfigure}{\linewidth}
        \includegraphics[width=\linewidth]{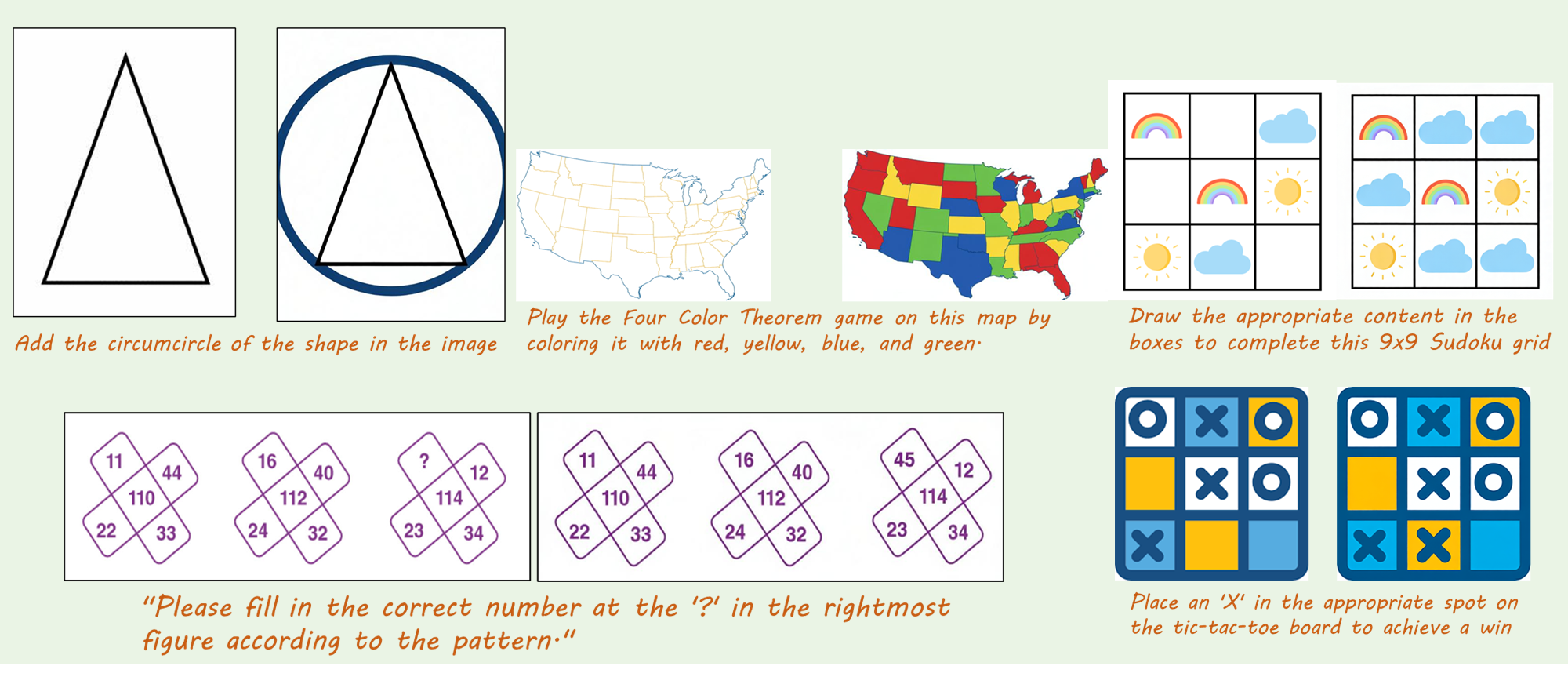}
    \end{subfigure}
    \begin{subfigure}{\linewidth}
        \includegraphics[width=\linewidth]{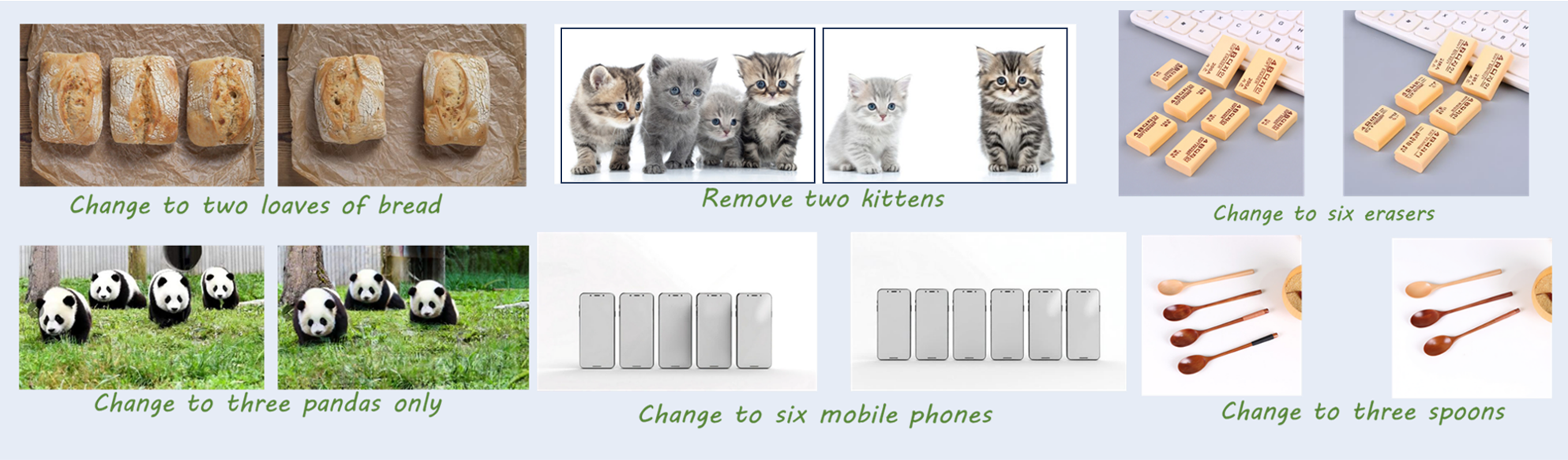}
    \end{subfigure}
    \caption{The editing results of our method on Kris-Bench.}
    \label{fig:kris_bench_results}
\end{figure*}
For the self-constructed COCO-based data, we employ a systematic approach to generate editing pairs. 
For each selected image, we apply instance segmentation to obtain object masks and select the largest segment as the target editing region based on mask area. 
We then create editing instructions following the template: ``Add a green segmentation mask (for object detection) to the \textit{[object name]} on the \textit{[position]} of the image'', where \textit{[object name]} and \textit{[position]} are automatically determined from the segmentation results.
This construction process yields paired data consisting of the original image, the edited image with the applied mask, and the corresponding text instruction.

\section{Knowledge-enhanced instruction} \label{app:knowledge-enhanced}

Complex editing instructions often contain implicit semantic requirements or require domain knowledge that is challenging for editing models to directly interpret. For instance, an instruction like ``Replace the sky with a sunset scene'' requires understanding what constitutes a visually plausible sunset (warm color gradients, sun position, atmospheric effects), while ``Change the car to a sports car'' demands knowledge of typical sports car characteristics (low profile, aerodynamic design, distinctive features). To bridge this semantic gap, we leverage Multi-modal Large Language Models (MLLMs) to decompose and enhance editing instructions with explicit, actionable guidance.

\begin{figure}[h]
    \centering
    \includegraphics[width=\linewidth]{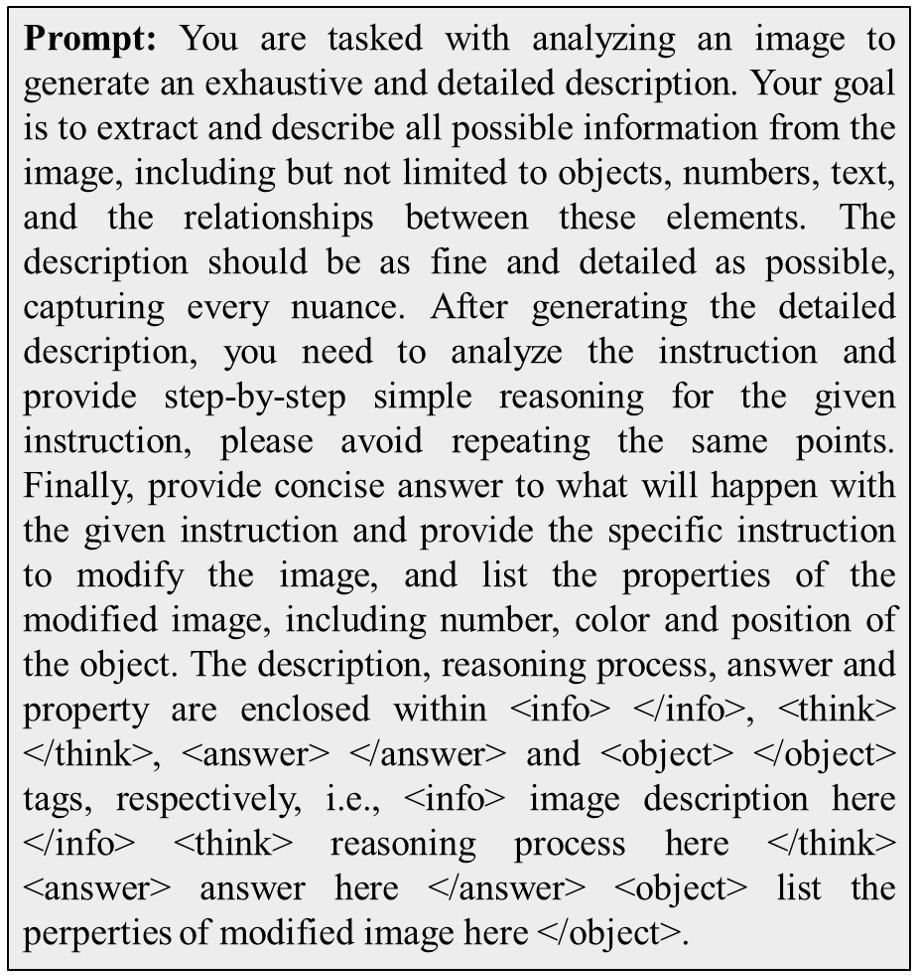}
    \caption{The template used for instruction enhancement.}
    \label{fig:knowledgeenhanced}
\end{figure}

\begin{figure*}[h]
    \centering
    \begin{subfigure}{\linewidth}
         \includegraphics[width=\linewidth]{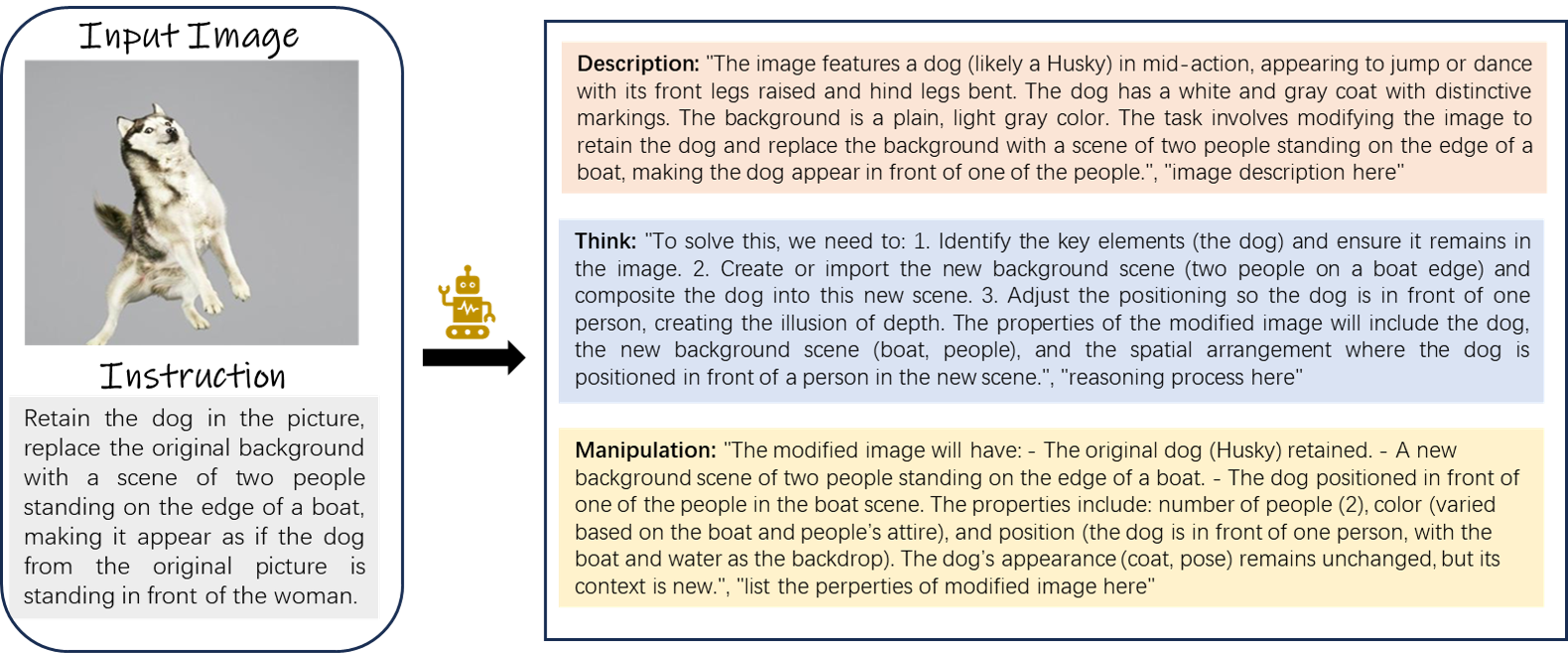}
         \caption{Sample 1.}
    \end{subfigure}
   \begin{subfigure}{\linewidth}
         \includegraphics[width=\linewidth]{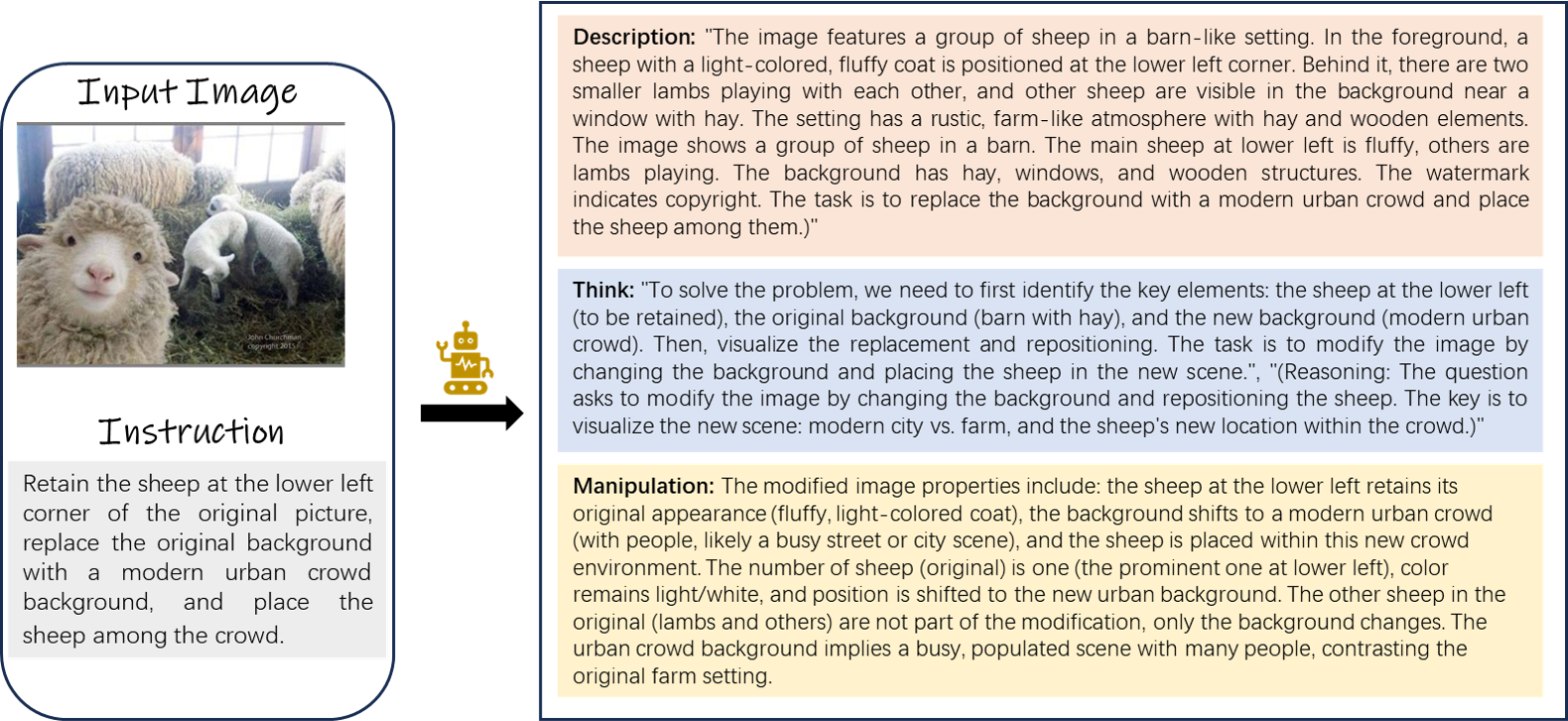}
          \caption{Sample 2.}
    \end{subfigure}
    \caption{Samples of MLLM-enhanced instruction.}
    \label{fig:knowledge-enhanced-instruction-samples}
\end{figure*}

\noindent\textbf{MLLM-based Instruction Enhancement.}
Given a source image and an editing instruction, we employ an MLLM (specifically, PeBR-R1-7b~\cite{chen2025perception}) to analyze the visual content and generate a knowledge-enhanced instruction. The enhancement process follows a structured prompt template that guides the MLLM to:
\begin{itemize}
    \item \textbf{Semantic Decomposition}: Break down abstract editing requests into specific visual attributes (colors, shapes, positions, textures).
    \item \textbf{Knowledge Injection}: Incorporate domain-specific knowledge relevant to the editing task (e.g., anatomical correctness, physical plausibility).
    \item \textbf{Spatial Grounding}: Provide explicit spatial references to guide precise localization of editing regions.
\end{itemize}
The used template is presented in Figure~\ref{fig:knowledgeenhanced}.
The MLLM processes both the image and instruction, generating enhanced instructions that explicitly specify editing details.
And we provide samples of the enhanced instructions in Figure~\ref{fig:knowledge-enhanced-instruction-samples}, including the description of the input image, the think process and the enhanced instruction.
The knowledge enhancement provides several key advantages. First, it transforms ambiguous instructions into explicit specifications, reducing the model's burden of implicit reasoning during the editing process. Second, it ensures semantic consistency by incorporating factual knowledge (e.g., correct number of petals for specific flowers, appropriate proportions for objects). Third, it provides fine-grained attribute specifications that align with our Dynamic Token Focus Relocation mechanism—the enhanced instructions contain more tokens describing precise attributes, which our mechanism can effectively emphasize during processing.

\section{Experiment}
\label{app:experiment}
\subsection{Reward Model} \label{app:reward_get}

Effective reward signals are critical for GRPO optimization. We design a comprehensive reward model based on Multi-modal Large Language Models (MLLMs) that evaluates edited images across multiple dimensions. Unlike single-metric rewards (e.g., CLIP score) that only measure text-image alignment, our reward model provides multi-faceted evaluation considering instruction following, visual coherence, and source consistency—essential criteria for high-quality image editing.

\noindent\textbf{MLLM-based Reward Evaluation.}
We employ Qwen2.5-VL-7B-Instruct as our reward model. Given an edited image, the original image, and the editing instruction, the model evaluates the editing quality through a structured prompt that assesses multiple aspects:

\begin{enumerate}
    \item \textbf{Word-level Analysis}: Extract key words related to subjects, objects, colors, numbers, lighting, style, and activities in the instruction, scoring how well each element is visually represented in the edited image. A special token \texttt{[No\_mistakes]} indicates whether all elements are correctly depicted.
    
    \item \textbf{Holistic Assessment}: Provide overall scores on four axes (each rated 1-5):
    \begin{itemize}
        \item \textit{Alignment Score}: How well the edited image matches the instruction in terms of content.
        \item \textit{Coherence Score}: Logical consistency of the image, including absence of visual glitches, object distortions, location errors, or incorrect colors/quantities.
        \item \textit{Style Score}: Aesthetic quality and visual appeal of the edited image.
        \item \textit{Consistency Score}: Fidelity to the original image in unedited regions.
    \end{itemize}
\end{enumerate}

The complete reward prompt is:

\begin{small}
\begin{quote}
\textit{
You are presented with an edited image (the first image) and its original image (the second image), and the associated text caption of the edited image. Your task is to analyze the edited image across multiple dimensions in relation to the caption and its original image. Specifically:}

\textit{1. Extract key words related to: subject, object, color, number, lighting, style and activities in the caption based on how well it is visually represented in the edited image.
   - A higher score indicates that the word is less well represented in the image.
   - The special token [No\_mistakes] represents whether all elements in the caption were correctly depicted. A high score suggests no mistakes; a low score suggests missing or incorrect elements.
}
\textit{2. Provide overall assessments for the image along the following axes (each rated from 1 to 5):
   - Alignment Score: How well the image matches the caption in terms of content.
   - Coherence Score: How logically consistent the image is (object location error, absence of visual glitches, object distortions, object is missing or extra, color or quantity error etc.).
   - Style Score: How aesthetically appealing the image looks, regardless of caption accuracy.
   - Consistency Score: How well the edited image consistent to its original image.}

\textit{Output your evaluation using the format below and also provide the reason why you give this score:
Alignment Score (1-5): X
Coherence Score (1-5): Y
Style Score (1-5): Z
Consistency Score (1-5): W
Output the basis and reasons for the scores given above.
}
\end{quote}
\end{small}

\begin{figure*}[h]
    \centering
    \includegraphics[width=\linewidth]{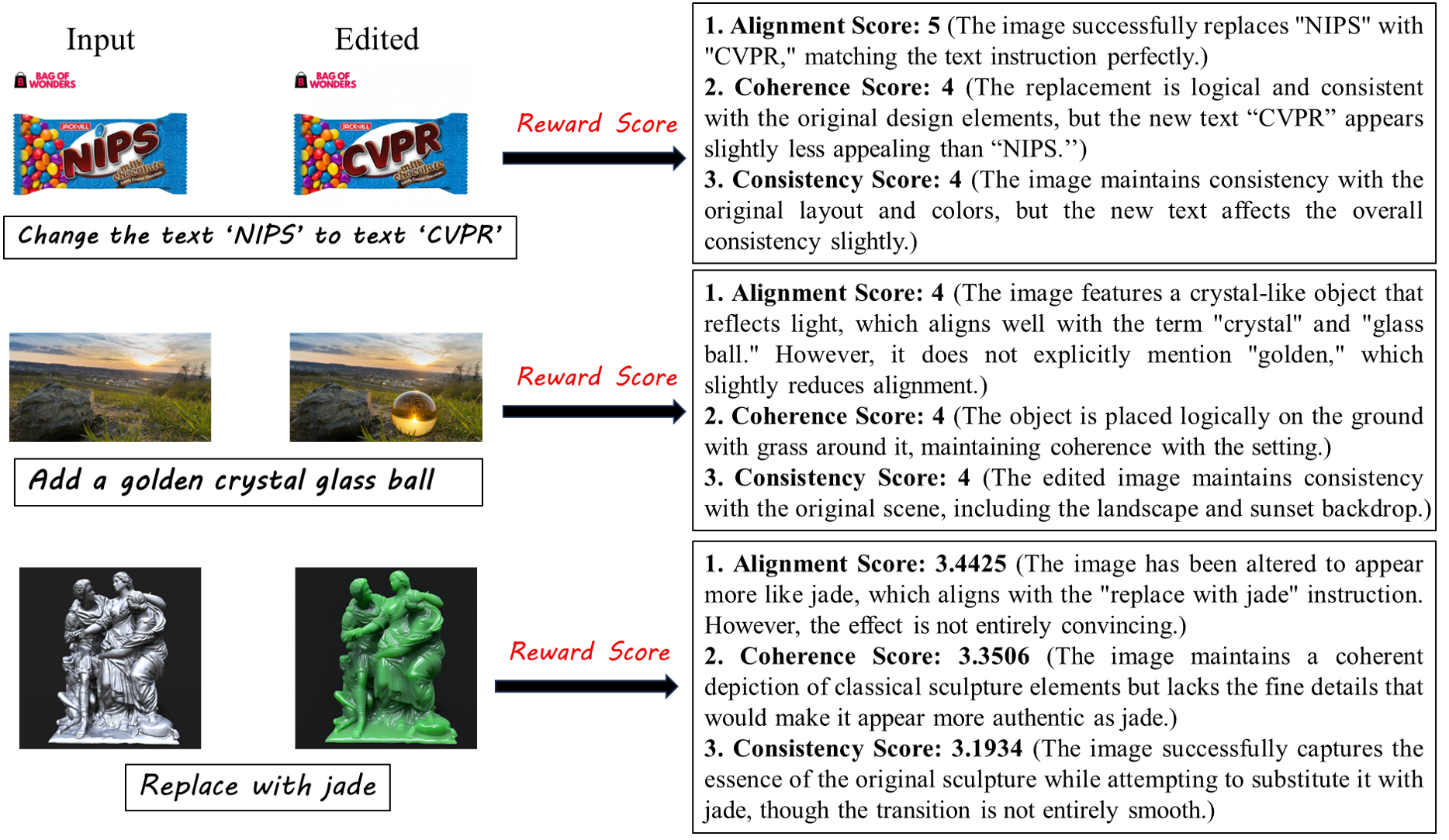}
    \caption{The samples of how the reward model evaluates the edited image.}
    \label{fig:reward_model_samples}
\end{figure*}
\noindent\textbf{Reward Computation.}
From the MLLM's structured output, we extract three key metrics to compute the final reward. We exclude the Style Score as it measures aesthetic appeal independent of editing correctness. The reward $R$ for each edited image is computed as:
\begin{equation}
R = \ S_{\text{alignment}} + S_{\text{coherence}} + S_{\text{consistency}},
\end{equation}
where $S_{\text{alignment}}$, $S_{\text{coherence}}$, and $S_{\text{consistency}}$ are the alignment, coherence, and consistency scores respectively, each normalized to [0, 5]. This formulation balances three critical aspects: (1) instruction-image alignment ensures the edit follows the textual specification, (2) coherence prevents visual artifacts and semantic inconsistencies, and (3) consistency preserves the integrity of unedited regions. By averaging these three scores, we obtain a holistic reward that guides the model to produce edits that are simultaneously accurate, plausible, and faithful to the source image. We provide the samples of how the reward model evaluates the edited image in Figure~\ref{fig:reward_model_samples}.

\noindent\textbf{Advantages of MLLM-based Rewards.}
Compared to traditional metrics like CLIP score or learned discriminators, our MLLM-based reward model offers several advantages. First, it provides interpretable, fine-grained feedback through word-level analysis and multi-dimensional scoring. Second, it can evaluate complex semantic correctness (e.g., correct number of objects, appropriate spatial relationships) beyond simple feature matching. Third, by conditioning on both the source and edited images, it explicitly evaluates source consistency—a crucial aspect often overlooked by generation-focused metrics. Empirically, we find that this reward formulation leads to more stable training and better alignment with human preferences compared to single-metric alternatives.

\subsection{Results on GEdit-Bench} \label{app:res_gedit}
\begin{figure*}[h]
    \centering
    \includegraphics[width=\linewidth]{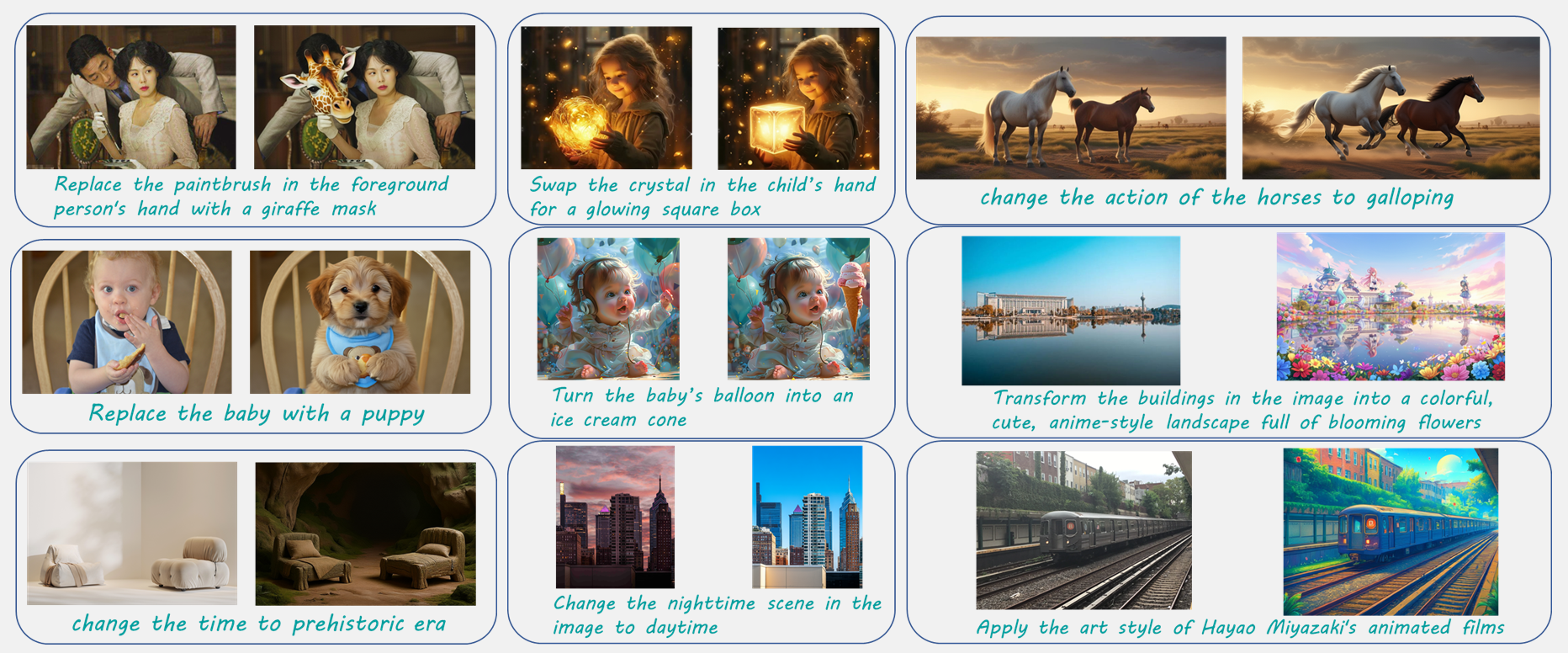}
    \caption{The editing results of our method on the GEdit-Bench.}
    \label{fig:gedit_bench_results}
\end{figure*}

\textbf{Editability Preservation.} Our dense reward optimization is specifically designed to enhance fine-grained alignment with detailed prompt features (e.g., precise colors, positions, quantities), which represents a different optimization objective than general-purpose editing. To assess whether this specialization compromises broader editing capabilities, we evaluate on GEdit-Bench with three metrics: Q\_SC (source consistency), Q\_PQ (edited quality), and Q\_O (overall quality), evaluated by QwenVL-2.5-72B. As shown in Table~\ref{tab:gedit_bench_results}, our method maintains competitive performance across both the intersection subset and full set, with scores comparable to top-performing general-purpose models like Qwen-Image and Step1X-Edit. While slight variations in individual metrics reflect the inherent trade-off between optimizing for fine-grained precision versus general editing flexibility, our method does not exhibit significant degradation despite being optimized for a more specialized objective. These results validate our design choice: we achieve substantial improvements on fine-grained tasks (as shown in Kris-Bench) without sacrificing competitiveness on general editing scenarios, demonstrating that dense reward optimization successfully balances fine-grained instruction following with general editability. 
\begin{table}[h]
    \centering
    \caption{The results on the GEdit-Bench,  where Q\_SC is the quality score of the source image, Q\_PQ is the quality score of the edited image, and Q\_O is the quality score of the edited image..}
     \resizebox{\linewidth}{!}{
    \begin{tabular}{c|ccc|ccc}
        \toprule
        \multirow{2}{*}{\textbf{Method}} & \multicolumn{3}{c|}{\textbf{Intersection subset} } & \multicolumn{3}{c}{\textbf{Full set}} \\
        ~ & Q\_SC & Q\_PQ & Q\_O& Q\_SC & Q\_PQ & Q\_O\\
        \midrule
        \textbf{InsPix2Pix} & 4.833 & 6.992 & 4.691& 4.746 & 6.913 & 4.578 \\
         \textbf{MagicBrush} & 5.814 & 7.149 & 5.653 & 5.752& 7.069& 5.558\\
          \textbf{AnyEdit} & 3.873 & 6.754 & 3.789 & 3.713 & 6.730 & 3.635\\
          \textbf{Step1X-Edit} & 7.501 & 7.264 & 7.189 & 7.388 & 7.279 & 7.067\\
          \textbf{OmniGen2} & 6.584 & 7.233 & 6.295 & 6.618 & 7.191 & 6.296 \\
          \midrule
          \textbf{Qwen-Image-Edit} & \textbf{7.819} & \underline{7.398} & \textbf{7.462} & \textbf{7.752} & \underline{7.394} & \textbf{7.402}\\
          \textbf{Ours} & \underline{7.723} & \textbf{7.441} & \underline{7.408} & \underline{7.618} & \textbf{7.417} & \underline{7.285}\\
        \bottomrule
    \end{tabular}
    }
    \label{tab:gedit_bench_results}
\end{table}
The editing results of our method on the GEdit-Bench are shown in Figure~\ref{fig:gedit_bench_results}.

\subsection{Visualization} \label{app:visual}
We provide qualitative comparisons to demonstrate \model's superiority in handling fine-grained instructions and knowledge-intensive editing tasks. The visualizations illustrate how our method achieves superior instruction following while maintaining visual quality and source consistency.

\textbf{Compare to Instruction-based methods.} Figure~\ref{fig:comparison_instruction_based} presents qualitative comparisons between \model and state-of-the-art instruction-based editing methods: Qwen-Image-Edit-r1, Step1X-Edit, and OmniGen2. These examples highlight scenarios requiring precise attribute control, including exact color specifications, numerical quantities, and spatial positioning.

The visual comparisons reveal critical differences in fine-grained instruction following capabilities. \model consistently captures precise specifications that baseline methods fail to reproduce. Step1X-Edit and OmniGen2 exhibit particular difficulty with numerical attributes and spatial positioning, often ignoring these fine-grained details in favor of general semantic understanding. Qwen-Image-Edit-r1 performs better but still shows inconsistent handling of multi-attribute instructions. These results validate that our Dense GRPO Optimization and Dynamic Token Focus Relocation mechanism effectively enhance fine-grained attribute control.

\noindent\textbf{Comparison with Knowledge-based Methods.}
Figure~\ref{fig:comparison_knowledge_based} compares \model against closed-source commercial VLM systems (Gemini 2, Doubao, GPT-4O) and chain-of-thought reasoning methods (Uni-CoT, UniWorld). These examples focus on knowledge-intensive editing tasks requiring domain expertise, factual knowledge, or complex reasoning about visual semantics.  We further provide the editing results of our method on Kris-Bench in Figure~\ref{fig:kris_bench_results}.

The knowledge-based comparisons highlight the importance of balancing semantic correctness with source fidelity. Commercial VLM systems like GPT-4O and Gemini 2 demonstrate strong instruction understanding and can generate semantically plausible content, but they often treat editing as a generation task, leading to poor source consistency—edited images may exhibit different lighting, style, or unnecessary modifications to unedited regions. Doubao shows similar issues with additional artifacts in complex scenes. Chain-of-thought methods (Uni-CoT, UniWorld) improve reasoning capabilities through explicit decomposition, but their editing quality remains inconsistent, particularly for instructions requiring both reasoning and precise visual control. 
In contrast, \model achieves superior balance: the MLLM-based reasoning component ensures semantic correctness and factual knowledge integration, while Dense GRPO Optimization maintains source consistency and visual quality. 

\subsection{User Study} \label{app:user_study}
To assess perceptual quality and editing fidelity from a human perspective, we conduct a comprehensive user study comparing \model against state-of-the-art instruction-based editing methods and commercial VLM systems. We recruit 10 participants with diverse backgrounds in computer vision and image editing. Each participant evaluates a series sampled editing pairs from Kris-Bench, ensuring broad coverage of knowledge-intensive scenarios.

\noindent\textbf{Evaluation Protocol.}
Participants rate each edited image on four criteria using a 5-point Likert scale (1=poor, 5=excellent):
\begin{itemize}
    \item \textbf{Editing Quality}: Visual plausibility, realism, and absence of artifacts in edited regions.
    \item \textbf{Instruction Adherence}: Alignment between the edited result and the textual instruction, including fine-grained attributes.
    \item \textbf{Consistency}: Preservation of unedited regions and coherence between edited and original content.
    \item \textbf{Overall Score}: Holistic assessment considering all aspects.
\end{itemize}
The evaluation is double-blind—participants are unaware of which method generated each result, and samples are presented in random order. For each editing task, we compute the mean score across all participants, and overall performance is obtained by averaging across all editing tasks. 

\textbf{For instruction-based editing methods.} 
Table~\ref{tab:user_study_instruction_based} compares \model against specialized instruction-based editing methods. Our approach achieves the highest scores across all criteria: editing quality (3.71), instruction adherence (3.49), consistency (3.60), and overall performance (3.47). Notably, \model outperforms the strong baseline Qwen-Image-Edit-r1 by significant margins—7.1\% in editing quality and 28.4\% in instruction adherence—demonstrating the effectiveness of dense GRPO optimization for fine-grained control. 
\begin{table}[h]
    \centering
    \caption{The results of the user study for instruction-based editing methods.}
     \resizebox{\linewidth}{!}{
    \begin{tabular}{c|cccc}
        \toprule
        \multirow{2}{*}{\textbf{Method}} & \multicolumn{3}{c}{\textbf{Editing Quality}} \\
        ~ & Quality & Adherence & Consistency & Overall \\
        \midrule
        Step1X-Edit & 2.743&	3.103	&2.713 &	2.790 \\
        OmniGen2 & 	2.903	& 1.984	& 3.040 &2.606\\
        Qwen-Image-Edit-r1 & 3.468	& 2.718	& 3.281 & 2.968\\
        Ours & \textbf{3.712} & \textbf{3.487}	& \textbf{3.603}	& \textbf{3.465}\\
        \bottomrule
    \end{tabular}
    }
    \label{tab:user_study_instruction_based}
\end{table}

%Step1X-Edit shows balanced but mediocre performance (2.74-3.10 range), likely due to its reliance on standard supervised learning without explicit reward optimization. OmniGen2 exhibits a critical weakness in instruction adherence (1.98), scoring nearly 50\% lower than our method, which suggests its difficulty in capturing precise textual specifications despite maintaining reasonable visual consistency (3.04). This validates our design hypothesis: trajectory-level gradient flow and dynamic token focus are essential for precise instruction following.

\textbf{For knowledge-based editing methods.} Table~\ref{tab:user_study_knowledge_based} compares \model against commercial VLM systems on knowledge-intensive editing tasks.

\begin{table}[h]
    \centering
    \caption{The results of the user study for knowledge-based editing methods.}
    \resizebox{\linewidth}{!}{
    \begin{tabular}{c|cccc}
        \toprule
        \multirow{2}{*}{\textbf{Method}} & \multicolumn{3}{c}{\textbf{Editing Quality}} \\
        ~ & Quality & Adherence & Consistency & Overall \\
        \midrule
        Gemini 2 & 3.021 & 3.387 & 	2.868 & 3.215	\\
        GPT-4o & 3.184 &	3.450 &	3.059 &	\textbf{3.503 }      \\
        Ours  & \textbf{3.712} & \textbf{3.48}7	& \textbf{3.603}	& 3.465\\
        \bottomrule
    \end{tabular}
    }
    \label{tab:user_study_knowledge_based}
\end{table}

\model significantly outperforms both GPT-4O and Gemini 2 in editing quality (3.71 vs. 3.18 and 3.02) and consistency (3.60 vs. 3.06 and 2.87). Interestingly, GPT-4o achieves the highest instruction adherence (3.45), slightly above our method, but suffers from poor consistency, revealing a fundamental trade-off in VLM-based editing: these systems excel at understanding instructions but struggle to maintain source image fidelity during edits, as they treat editing more like generation tasks. Gemini 2's particularly weak consistency suggests even stronger generation bias. In contrast, \model achieves balanced performance across all metrics, validating that MLLM-based reasoning combined with dense reward optimization successfully preserves source consistency while following complex instructions. The overall scores (3.47 for \model vs. 3.50 for GPT-4o) are competitive, but our superior editing quality and consistency make \model more suitable for practical editing applications where source preservation is critical.

\end{document}